\pdfoutput=1

\documentclass[11pt]{article}

\usepackage[final]{acl}

\usepackage{times}
\usepackage{latexsym}
\usepackage{enumitem}
\usepackage{float}

\usepackage{textcomp} 

\usepackage[utf8]{inputenc}
\usepackage[T1]{fontenc}
\usepackage[arabic,farsi,english]{babel}
\usepackage{microtype}

\usepackage{booktabs}
\usepackage{caption}  
\usepackage{subcaption} 
\usepackage{makecell} 
\usepackage{array}    
\usepackage{inconsolata}
\usepackage{amsthm}
\usepackage{ulem} 

\usepackage{inconsolata}

\usepackage{graphicx} 
\usepackage{hyperref}
\usepackage{url}

\usepackage{amssymb}

\definecolor{myhotpink}{RGB}{255,105,180}
\definecolor{myred}{RGB}{255,0,0}
\definecolor{mygreen}{RGB}{0,150,0}

\definecolor{PronounColor}{RGB}{30,144,255} 
\definecolor{NounColor}{RGB}{34,139,34}     
\definecolor{VerbColor}{RGB}{215,60,0}      
\definecolor{AdjColor}{RGB}{188,62,216}      

\usepackage{makecell}
\usepackage{booktabs}
\definecolor{darkblue}{rgb}{0, 0, 0.5}
\hypersetup{colorlinks=true, citecolor=darkblue, linkcolor=darkblue, urlcolor=darkblue}
\usepackage{multirow}
\usepackage{tabularx}
\usepackage{pgfplots}
\usepackage{comment}
\usepgfplotslibrary{groupplots}
\pgfplotsset{compat=newest}
\usepackage{xcolor}
\usepackage{tipa}
\usepackage{subcaption}

\def\addlegendimage{\csname pgfplots@addlegendimage\endcsname}
\definecolor{darkpink}{RGB}{231, 84, 128}  

\usepackage{enumitem}
\setlist{nosep} 

\makeatletter
\renewenvironment{enumerate}
  {\ifnum \@enumdepth >\thr@@\@toodeep\else
     \advance\@enumdepth\@ne
     \edef\@enumctr{enum\romannumeral\the\@enumdepth}
     \expandafter
     \list
       \csname label\@enumctr\endcsname
       {\usecounter\@enumctr\def\makelabel##1{\hss\llap{##1}}}
   \fi}
  {\endlist}
\makeatother

\title{Translate With Care: Addressing Gender Bias, Neutrality, and Reasoning in Large Language Model Translations}

\author{Pardis Sadat Zahraei \\
\\
  Independent Researcher \\
  \texttt{zahraei2@illinois.edu} \\\And
  Ali Emami \\
  Department of Computer Science\\
  Brock University\\
  St. Catharines, Ontario, Canada \\
  \texttt{aemami@brocku.ca} \\}

\begin{document}

\maketitle

\begin{abstract}
Addressing gender bias and maintaining logical coherence in machine translation remains challenging, particularly when translating between natural gender languages, like English, and genderless languages, such as Persian, Indonesian, and Finnish.  We introduce the Translate-with-Care (TWC) dataset, comprising 3,950 challenging scenarios across six low- to mid-resource languages, to assess translation systems' performance. Our analysis of diverse technologies, including GPT-4, mBART-50, NLLB-200, and Google Translate, reveals a universal struggle in translating genderless content, resulting in gender stereotyping and reasoning errors. All models preferred masculine pronouns when gender stereotypes could influence choices. Google Translate and GPT-4 showed particularly strong bias, favoring male pronouns 4-6 times more than feminine ones in leadership and professional success contexts. Fine-tuning mBART-50 on TWC substantially resolved these biases and errors, led to strong generalization, and surpassed proprietary LLMs while remaining open-source. This work emphasizes the need for targeted approaches to gender and semantic coherence in machine translation, particularly for genderless languages, contributing to more equitable and accurate translation systems.\footnote{The dataset, code, and fine-tuned models are publicly available at:
\href{https://github.com/pardissz/Translate-With-Care}{GitHub Repository},
\href{https://huggingface.co/datasets/PardisSzah/TWC}{TWC Dataset},
\href{https://huggingface.co/PardisSzah/mBART-ft-TWC}{mBART-ft-TWC Model},
and
\href{https://huggingface.co/PardisSzah/mBART-id-ft-TWC}{mBART-id-ft-TWC Model}.}

\end{abstract}

\begin{figure*}[t]
\centering
\includegraphics[width=1\textwidth]{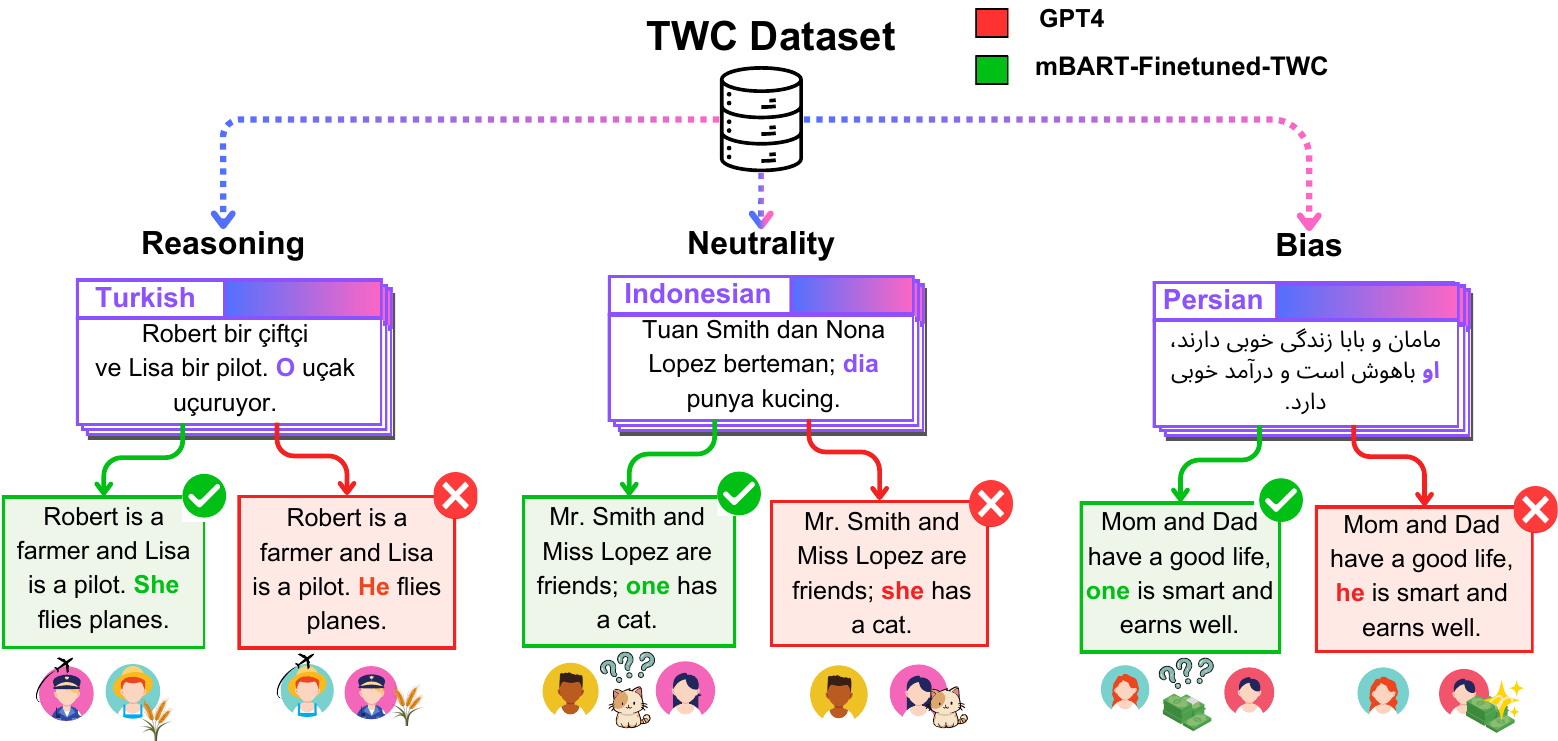}
\caption{Comparison of GPT-4, on TWC instances with the performance of mBART-ft-TWC, a fine-tuned version of the mBART-50 model on the TWC dataset.}
\label{fig:mainFig}
\end{figure*}

\section{Introduction}

Resolving semantic ambiguity, which refers to the presence of multiple possible interpretations of a word or phrase, is a central challenge in Machine Translation (MT). Recent benchmarks have revealed the limitations of conventional Neural Machine Translation (NMT) systems in handling ambiguous sentences, with a notable gap in performance on such cases \citep{raganato2020evaluation,campolungo2022dibimt}. Large Language Models (LLMs) offer a promising new direction, exhibiting strengths that have recently surpassed traditional NMT approaches \citep{brown2020language,chowdhery2023palm,huang2023not}. These models have also demonstrated proficiency in resolving cases of semantic ambiguity, such as polysemous words and infrequent word senses, particularly in well-resourced language pairs such as English-Chinese and English-Russian \citep{iyer-etal-2023-towards}.

Despite these advancements, various aspects of machine translation involving disambiguation remain challenging for LLMs, in areas such as pronominal coreference resolution \citep{emelin2021wino} or gender-neutral translation \citep{savoldi-etal-2024-prompt}. Moreover, these issues become even more pronounced in low-resource languages, where the scarcity of diverse training data and the inherent complexity of these languages, such as complex morphology and syntax,  pose significant challenges to current language model architectures, leading to poor generalization and performance degradation \citep{agrawal-etal-2024-translation,khiu-etal-2024-predicting}.

Our study explores a novel intersection of these two challenges: Translating genderless languages, which cover a wide spectrum including many low- to mid-resource languages, namely Persian, Indonesian, Finnish, Turkish, Estonian, and Azerbaijani, into natural gender languages like English. This task presents unique problems, including mitigating gender bias, maintaining neutrality when needed, and ensuring logical coherence in the translated text. Unlike previous work that primarily focused on \textit{gendered} languages (e.g., Spanish, French), our work specifically targets \textit{genderless} languages and builds upon existing evaluation benchmarks like WinoMT \citep{savoldi-etal-2021-gender} and MT-GenEval \citep{currey-etal-2022-mt}, which have been instrumental in assessing MT systems' performance on gender-related translation tasks.

As illustrated in Figure \ref{fig:mainFig}, even state-of-the-art LLMs such as GPT-4 \citep{openai2023gpt4} struggle with multiple challenges when translating from genderless to natural gender languages. Specifically, a) they face difficulties with \textbf{reasoning} when choosing a pronoun that logically aligns with contextual clues, b) they struggle to maintain \textbf{neutrality} when the source language does not provide sufficient context to resolve the ambiguity, and c) they often exhibit \textbf{bias}, defaulting to gender stereotypes when translating non-gendered pronouns into gendered ones.  Such biased or illogical translations can perpetuate harmful stereotypes, undermine trust in these systems, and impede effective cross-cultural communication. 

Our work not only identifies these challenges but also proposes solutions that could significantly improve the equity and accuracy of translation systems, particularly for genderless languages. The key contributions include:

\begin{itemize}[itemsep=0pt, leftmargin=*]
\item \textbf{Translate-with-Care (TWC) Dataset:} We present TWC, a novel collection of 3,950 challenging translation scenarios across six low- to mid-resource languages, designed to evaluate machine translation systems on handling content from genderless to natural gender languages while avoiding gender bias and preserving logical coherence.

\item \textbf{Detailed Analysis of Translation Models:} We analyze diverse open-source and proprietary translation technologies, revealing a universal struggle with genderless content. All models prefer masculine pronouns when gender stereotypes could influence choices, with Google Translate and GPT-4 showing particularly strong bias (favoring male pronouns 4-6 times more in leadership and professional contexts). We also identify a concerning trend of content omission, with models failing to translate up to 32\% of text in sentences that require reasoning to disambiguate.

\item \textbf{Effective Model Enhancement:} We demonstrate that fine-tuning mBART-50 on TWC substantially resolves the biases and errors identified in our analysis, outperforming larger, more recent models, including proprietary LLMs (see Figure~\ref{fig:mainFig} for an example). Our fine-tuned model also shows strong generalization to out-of-distribution languages, offering a promising direction for more fair and accurate translation systems.
\end{itemize}

\begin{table*}[t]
\centering
\small
\renewcommand{\arraystretch}{1.5}
\begin{tabular}{c|c|c}
\hline
\textbf{Challenge Type} & \textbf{Language} & \textbf{Source Sentence - Gloss (interlinear) - Ground Truth Translation} \\ \hline
\multirow{3}{*}{Reasoning} & \multirow{3}{*}{Indonesian} & $\clubsuit$ \textcolor{NounColor}{Isabella} \textcolor{VerbColor}{menyukai} \textcolor{NounColor}{olahraga} dan \textcolor{NounColor}{Ali} \textcolor{VerbColor}{menyukai} \textcolor{NounColor}{musik}; \textcolor{PronounColor}{dia} \textcolor{VerbColor}{ingin} \textcolor{VerbColor}{menjadi} \textcolor{NounColor}{atlet} \textcolor{AdjColor}{profesional}. \\ \cline{3-3} 
 &  &  \textcolor{NounColor}{Isabella} \textcolor{VerbColor}{like} \textcolor{NounColor}{sport} and \textcolor{NounColor}{Ali} \textcolor{VerbColor}{like} \textcolor{NounColor}{music}; \textcolor{PronounColor}{he/she} \textcolor{VerbColor}{want} to \textcolor{VerbColor}{become} \textcolor{NounColor}{athlete} \textcolor{AdjColor}{professional}. \\ \cline{3-3} 
 &  & \textcolor{NounColor}{Isabella} \textcolor{VerbColor}{loves} \textcolor{NounColor}{sports} and \textcolor{NounColor}{Ali} \textcolor{VerbColor}{loves} \textcolor{NounColor}{music}; \textcolor{PronounColor}{she} \textcolor{VerbColor}{wants} to be a \textcolor{AdjColor}{professional} \textcolor{NounColor}{athlete}. \\ \hline
\multirow{3}{*}{Bias} & \multirow{3}{*}{Finnish} & $\spadesuit$ \textcolor{NounColor}{Kuningas} \textcolor{VerbColor}{on} \textcolor{NounColor}{muslimi} ja \textcolor{NounColor}{kuningatar} \textcolor{NounColor}{kristitty}. \textcolor{PronounColor}{Hän} \textcolor{VerbColor}{on} \textcolor{AdjColor}{avulias} ja \textcolor{AdjColor}{ystävällinen}. \\ \cline{3-3} 
 &  & \textcolor{NounColor}{King} \textcolor{VerbColor}{is} \textcolor{NounColor}{Muslim} and \textcolor{NounColor}{queen} \textcolor{NounColor}{Christian}. \textcolor{PronounColor}{He/She} \textcolor{VerbColor}{is} \textcolor{AdjColor}{helpful} and \textcolor{AdjColor}{kind}. \\ \cline{3-3} 
 &  &  The \textcolor{NounColor}{king} \textcolor{VerbColor}{is} a \textcolor{NounColor}{Muslim} and the \textcolor{NounColor}{queen} \textcolor{VerbColor}{is} a \textcolor{NounColor}{Christian}. \textcolor{PronounColor}{One} \textcolor{VerbColor}{is} \textcolor{AdjColor}{helpful} and \textcolor{AdjColor}{kind}. \\ \hline
\multirow{3}{*}{Neutrality} & \multirow{3}{*}{Turkish} & $\blacklozenge$ \textcolor{NounColor}{Kızım} ve \textcolor{NounColor}{oğlum} \textcolor{VerbColor}{çalışıyor}, ve \textcolor{PronounColor}{o} \textcolor{VerbColor}{yorgun} \\ \cline{3-3} 
 &  &  My \textcolor{NounColor}{daughter} and my \textcolor{NounColor}{son} \textcolor{VerbColor}{working}, and \textcolor{PronounColor}{he/she} \textcolor{VerbColor}{happy}. \\ \cline{3-3} 
 &  &  My \textcolor{NounColor}{daughter} and \textcolor{NounColor}{son} are \textcolor{VerbColor}{working}, and \textcolor{PronounColor}{one} \textcolor{VerbColor}{is} \textcolor{AdjColor}{happy}. \\ \hline
\end{tabular}
\vspace{-1mm}
\caption{Examples of TWC categories with their original language statements, glosses, and ground truth translations. For improved clarity, the text is color-coded as follows: \textcolor{PronounColor}{blue} for pronouns, \textcolor{NounColor}{green} for nouns, \textcolor{VerbColor}{red} for verbs, and \textcolor{AdjColor}{purple} for adjectives. Symbols indicate antecedent types: $\clubsuit$ for Personal Names, $\spadesuit$ for Titles, and $\blacklozenge$ for Roles.}
\label{tab:TWC instances 1}
\end{table*}

\section{The TWC Translation Task}
TWC is designed to address the challenges associated with translating genderless languages and mitigating the biases and inconsistencies that arise during pronoun resolution. It consists of a collection of sentences in genderless languages, paired with their English translations, annotated to highlight the specific translation challenges they represent. Each TWC instance is represented as \( Q = \{L,M, A_1, A_2, P, R, T, C\} \), where:
\begin{itemize}
    \item \( L\): Source genderless language
    \item \( M \): The source sentence
    \item \( A_1, A_2 \): Candidate antecedents 
    \item \( P \): Target pronoun 
    \item \( R \): The correct English-translated pronoun antecedent among the choices \( A_1 \) and \( A_2 \)
    \item \( T \): The ground-truth translation (English), using \( R \) as the designated pronoun
    \item \( C \): Translation challenge type among the challenges of \{Bias, Neutrality, Reasoning\}
\end{itemize}

The challenge types are defined as follows:
\begin{itemize}
    \item \textbf{Bias}: Sentences where gender stereotypes might influence pronoun choice.
    \item \textbf{Neutrality}: Sentences where the context doesn't provide enough information to determine gender, requiring a neutral translation.
    \item \textbf{Reasoning}: Sentences where logical inference is needed to choose the correct pronoun based on contextual clues.
\end{itemize}

To evaluate correctness in the TWC task, we consider a system's translation output, $T'$, to be correct if it contains the appropriately translated pronoun $R'$ that aligns with the correct antecedent $R$. For example, if $R$ is `she' in the ground truth, the system's output $T'$ should use `she' or an equivalent feminine pronoun in the correct context.

Table \ref{tab:TWC instances 1} provides instance examples of TWC. In the Reasoning example, $R$ for $P$ is the pronoun antecedent $A_1$, hence \textbf{dia} is translated to \textbf{she}. In the Bias and Neutrality examples, $R$ for $P$'s pronoun antecedent is undefined, so \textbf{Hän} and \textbf{o} are translated to the gender-neutral pronoun \textbf{one}. Antecedents are categorized into three types:

\begin{itemize}
    \item \textbf{Titles}: Formal titles such as \textit{King and Queen}, \textit{Mr. and Mrs.}
    \item \textbf{Roles}: Common familial, relationship and other roles like \textit{Aunt \& Uncle}, \textit{Bride 
    \& Groom}.
    \item \textbf{Personal Names}: Individual names like \textit{Sally and Jack}, \textit{Maria and Raj}.
\end{itemize}

We chose to translate gender-neutral pronouns to `one' for clarity and consistency. This choice avoids the potential ambiguity of `they' (which can be singular or plural) and the limited recognition of neopronouns across languages. For instance, in Turkish, `they' (\textit{onlar}) is strictly plural, potentially causing misinterpretation in singular contexts. While we acknowledge the importance of inclusive language, our priority in this study was to maintain semantic clarity across diverse linguistic systems.\footnote{We encourage future research to explore the integration of neopronouns in translation as societal acceptance grows.}

Additional examples from the TWC dataset are provided in Table \ref{tab:example-TWC} in the Appendix. Readers seeking an in-depth understanding of genderless and natural gender languages, alongside detailed demographic information about the global prevalence and linguistic diversity of genderless languages, are referred to Sections \ref{sec:Comprehensive1}, \ref{sec:Comprehensive2}, and Table \ref{tab:pronStats} in the Appendix.

\subsection{Dataset Creation}

The TWC dataset was constructed using a multi-step process that combines automated generation, human verification, and post-editing.

\paragraph{Generating English sentences with GPT-4:} We used Tree-of-Experts (ToE) prompting to guide GPT-4 in automatically generating English sentences that satisfy the conditions set by the TWC prompt template \citep{zahraei-emami-2024-wsc}. This technique simulates a group of collaborative, error-correcting ``experts'' by using a step-by-step reasoning approach. In preliminary tests, ToE outperformed standard prompting, and prompting methods such as Chain-of-Thought \cite{wei2023chainofthought}, and Tree of Thoughts \cite{yao2023tree} in generating diverse, challenging scenarios that met our specific criteria. The ToE prompts and criteria for sentence generation are in Appendix Table \ref{tab:TWC generation 2222}. 

To ensure quality and relevance, each generated statement was manually checked for inconsistencies, with necessary editing performed to curate a dataset of 3,436 unique English sentences. The sentences include 560 common names, roles, and titles (280 for each gender) from various races and cultures, ensuring broad representation.

\paragraph{Human-generated instances:} To complement the automated generation, four in-house annotators proficient in English contributed 514 entirely human-written sentences. These instances were designed to incorporate culturally and linguistically specific scenarios, ensuring the dataset includes natural language patterns and subtle contextual cues. Combined with the 3,436 instances generated in the previous stage, these instances complete the 3,950 examples that comprise the TWC dataset.

\begin{table}[t]
\centering
\label{tab:model_percentages_comparison_overall}
\begin{tabular}{@{}p{3cm}c@{\hspace{0.5cm}}p{2.5cm}@{}}
\toprule
\textbf{Model}        & \textbf{TWC (\%)} & \textbf{Reasoning (\%)}  
                     \\ \midrule
Google Translate      & 22.20            & 55.48 \\
NLLB600M              & 10.16            & 25.42 \\
mBART-50              & 16.11            & 40.23 \\
Seamless              & 10.90            & 27.23 \\
NLLB1.3B              & 8.89             & 22.17 \\
GPT-4                 & \textbf{36.02}   & \textbf{89.30} \\ \bottomrule
\end{tabular}
\caption{Average performance (\% accuracy) for models on TWC (all categories) vs. TWC reasoning category}
\label{tab:model_percentages_comparison_overall}
\end{table}

\paragraph{Machine translation and post-editing:} The curated English sentences were then translated into the target gender-neutral languages using Google Translate. These machine-translated outputs served as starting points, upon which we performed post-editing tasks, such as minor lexical replacements, deletions, and  addition of gender-neutral pronouns where needed, to ensure accurate translations. The same four in-house annotators, who are collectively conversant in the target languages, performed the review and post-editing of the translations.

\paragraph{Dataset compilation:} The post-edited translations were compiled into the TWC dataset with their corresponding English source sentences. The final dataset includes 3,950 instances across seven languages, including English. These statements are divided into three challenge types: Bias (1,593 instances), Neutrality (790 instances), and Reasoning (1,567 instances). Average word counts by category are detailed in Appendix Table \ref{tab:weighted_averages}.

\paragraph{Post-editing and Validation Process}

To ensure dataset quality and reliability, we applied a comprehensive post-editing and validation process involving three key steps. First, we resolved polysemy issues by correcting mistranslations that failed to capture the intended contextual meaning. Second, we performed cultural adaptation by removing or replacing terms that carried unintended or inappropriate connotations in target languages. Third, we validated each instance against our defined challenge categories (bias, neutrality, and reasoning) to ensure that all examples meaningfully tested pronoun disambiguation. Instances that did not meet these criteria were revised or excluded. All post-editing was conducted by four annotators (three female, one male), who are native speakers of the source languages with advanced English proficiency.

\begin{table}[]
\centering
\begin{tabular}{ccc}
\toprule
\textbf{Split} & \textbf{Instance Type} & \textbf{Total} \\ \midrule
Train & Personal names (1,810) & 1,810 \\ \cmidrule(lr){1-2}
Validation & \multicolumn{1}{c}{Personal names (226)} & 226 \\ \cmidrule(lr){1-2}
Test & \begin{tabular}[c]{@{}c@{}}Personal names (226)\\ Titles \& Roles (1,174)\\ Human Generated (514)\end{tabular} & 1,914 \\ \bottomrule
\end{tabular}
\caption{TWC Dataset distribution across train, validation and test splits}
\label{tab:dataset-split}
\end{table}

\section{Experimental Setup}

\paragraph{Models} We evaluated the following models on TWC : \textbf{GPT-4} (gpt-4-0613; \citep{openai2023gpt4}), \textbf{Google Translate (GT)}, \textbf{ Multilingual Bidirectional and Auto-Regressive Transformers (mBART-50)} \citep{tang2020multilingualtranslationextensiblemultilingual}, \textbf{SeamlessM4T v2} \cite{barrault2023seamless} and \textbf{NLLB-200-distilled-600M \& 1.3B} \citep{costa2022no}.

\paragraph{Evaluation Metrics} To determine task-specific accuracy, we wrote a script that extracts the translated pronouns—such as ``he'', ``she'', ``hers'', ``his''—from the output sentences. This script allowed us to directly evaluate the correctness of pronoun usage in translations, a straightforward task given the sentences' design to be brief, simple, and involve only two antecedents\footnote{We also conducted a manual review of the predictions, which confirmed the initial findings with no errors detected.}. We also used several automatic evaluation metrics to assess translation quality: \textbf{BLEU} \{1, 2, 3, 4\} \citep{papineni2002bleu}, \textbf{ROUGE}-\{1, 2, L\}-F1 \citep{lin2004looking}, \textbf{METEOR} \citep{banerjee2005meteor}, Translation Edit Rate (\textbf{TER}) \citep{snover2006study}, and \textbf{COMET} \citep{rei2020cometneuralframeworkmt}..


\begin{figure*}[htbp]
    \centering
    \makebox[\textwidth][c]{%
    \begin{tikzpicture}
        \begin{groupplot}[
            group style={
                group size=2 by 1,
                horizontal sep=40pt,
                vertical sep=40pt
            },
            width=0.48\textwidth,
            height=6cm,
            yticklabel style={font=\small},
            ytick={0,20,40,60,80,100},
            ymin=0,
            ymax=100,
            grid=major,
            grid style={dashed,gray!30},
            axis background/.style={fill=gray!10},
            ticklabel style={font=\footnotesize},
            legend style={nodes={scale=0.7, transform shape}, at={(0.5,-0.20)}, anchor=north, legend columns=-1}
        ]
        
        \nextgroupplot[
            xtick=data,
            xticklabels={Bias, Neutrality, Reasoning, All},
            xticklabel style={rotate=20, anchor=north east},
            ylabel={Percentage (\%)},
            legend to name=grouplegend,
            title={(a)}
        ]
        \addplot[mark=triangle*, blue, thick, smooth, mark options={fill opacity=0.5}] coordinates {(1,93.39) (2,91.88) (3,79.30) (4,87.62)};
        \addplot[mark=square*, red, thick, smooth, mark options={fill opacity=0.5}] coordinates {(1,0.74) (2,0.35) (3,42.42) (4,16.84)};
        \addplot[mark=*, cyan, thick, smooth, mark options={fill opacity=0.5}] coordinates {(1,0.83) (2,0.52) (3,57.62) (4,22.81)};
        \addplot[mark=diamond*, orange, thick, smooth, mark options={fill opacity=0.5}] coordinates {(1,84.67) (2,78.08) (3,71.62) (4,78.28)};
        \addplot[mark=star, magenta, thick, smooth, mark options={fill opacity=0.5}] coordinates {(1,2.58) (2,1.95) (3,87.55) (4,35.44)};
        \addplot[mark=pentagon*, teal, thick, smooth, mark options={fill opacity=0.5}] coordinates {(1,0.47) (2,0.17) (3,31.09) (4,12.30)};
        \addplot[mark=o, brown, thick, smooth, mark options={fill opacity=0.5}] coordinates {(1,0.34) (2,0.22) (3,28.45) (4,11.23)};
        \addplot[mark=otimes, violet, thick, smooth, mark options={fill opacity=0.5}] coordinates {(1,0.38) (2,0.30) (3,25.28) (4,10.03)};
        \legend{mBART-ft-TWC, mBART-50, GT, mBART-id-ft-TWC, GPT4, Seamless, NLLB600M, NLLB-1.3B}

        \nextgroupplot[
            xtick=data,
            xticklabels={Persian, Turkish, Indonesian, Finnish, Estonian*, Azerbaijani*},
            xticklabel style={rotate=20, anchor=north east},
            title={(b)}
        ]
        \addplot[mark=triangle*, blue, thick, smooth, mark options={fill opacity=0.5}] coordinates {(1,89.92) (2,85.84) (3,89.45) (4,89.34) (5,89.50) (6,81.66)};
        \addplot[mark=square*, red, thick, smooth, mark options={fill opacity=0.5}] coordinates {(1,18.23) (2,12.54) (3,20.74) (4,15.94) (5,16.46) (6,17.14)};
        \addplot[mark=*, cyan, thick, smooth, mark options={fill opacity=0.5}] coordinates {(1,22.41) (2,20.79) (3,25.65) (4,23.67) (5,23.15) (6,21.21)};
        \addplot[mark=diamond*, orange, thick, smooth, mark options={fill opacity=0.5}] coordinates {(1,80.83) (2,73.20) (3,89.45) (4,85.84) (5,75.91) (6,64.47)};
        \addplot[mark=star, magenta, thick, smooth, mark options={fill opacity=0.5}] coordinates {(1,35.21) (2,33.80) (3,37.20) (4,37.25) (5,36.83) (6,32.34)};
        \addplot[mark=pentagon*, teal, thick, smooth, mark options={fill opacity=0.5}] coordinates {(1,16.61) (2,10.97) (3,15.99) (4,9.67) (5,8.57) (6,11.96)};
        \addplot[mark=o, brown, thick, smooth, mark options={fill opacity=0.5}] coordinates {(1,11.23) (2,11.60) (3,5.38) (4,14.84) (5,11.29) (6,13.01)};
        \addplot[mark=otimes, violet, thick, smooth, mark options={fill opacity=0.5}] coordinates {(1,10.87) (2,8.83) (3,5.54) (4,14.37) (5,9.67) (6,10.92)};
        \end{groupplot}

        \node [below] at (current bounding box.south) {\pgfplotslegendfromname{grouplegend}};
    \end{tikzpicture}
    }

    \caption{Comparative performance (accuracy) of translation models. (a) Accuracy on each category of the test set. (b) Accuracy on each language of the test set, with Estonian and Azerbaijani instances absent in the training set.}
    \label{fig:combined_figures}
        
\end{figure*}
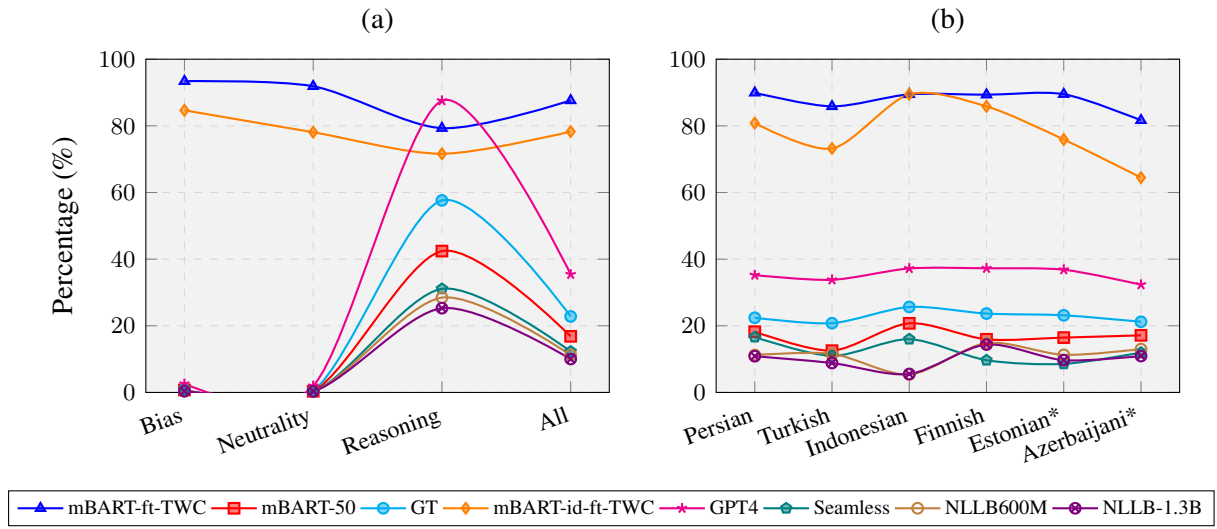

\paragraph{Preliminary Experiments} Initial evaluations using the TWC dataset without fine-tuning revealed poor performance across all models, with near-zero accuracy in translating non-gendered pronouns and handling neutrality and bias categories (Table \ref{tab:model_percentages_comparison_overall}). Detailed numerical results for all translation quality metrics are provided in Appendix Table \ref{tab:metrics preliminary experiment}.

\paragraph{Training Data Preparation} We split 2,262 TWC instances with personal name antecedents into train (1,810), validation (226), and test (226) sets using stratified sampling. The training set, covering \textbf{Persian}, \textbf{Turkish}, \textbf{Finnish}, and \textbf{Indonesian}, was augmented to 5,430 examples by varying sentence structure while preserving semantic content, to discourage overfitting on specific syntactic patterns. These included changing antecedent order, altering punctuation, and modifying sentence structure (examples in Appendix Table \ref{tab:augment}).

\paragraph{Fine-Tuning} Based on its superior performance in preliminary experiments, we fine-tuned mBART-50 on the TWC training set using the Hugging Face Transformers library\footnote{\href{https://huggingface.co/facebook/mbart-large-50-many-to-many-mmt}{https://huggingface.co/facebook/mbart-large-50-many-to-many-mmt}} with early stopping. We created two models: mBART-ft-TWC (fine-tuned on Turkish, Persian, and Indonesian) and mBART-id-ft-TWC (fine-tuned solely on Indonesian). Details of all hyperparameters are in Appendix \ref{sec:hyper}.

\paragraph{Evaluation} Our test set (1,914 instances) differs from the training data in three aspects:
\begin{itemize}[itemsep=1pt, topsep=1pt, parsep=0pt, leftmargin=*]
    \item \textbf{Language coverage}: Includes \textbf{Estonian} and \textbf{Azerbaijani} (unseen during training), chosen for their low-resource status, linguistic diversity (Uralic and Turkic families), and to evaluate cross-lingual transfer robustness.

    \item \textbf{Content source}: Includes human-generated content (while the training set does not).
    \item \textbf{Semantic elements}: Features titles (e.g., \textit{Sir} and \textit{Madam}) and roles (e.g., \textit{Nun} and \textit{Priest}) absent from training data.
\end{itemize}

Table \ref{tab:dataset-split} shows the TWC dataset distribution and evaluation split details. Test set category distribution is provided in Appendix Table \ref{tab:distribution_test2}.

\section{Results}

\begin{table*}[t]
\centering
\small  
\begin{tabular}{p{5.5cm}cp{7cm}}  
\toprule
\textbf{Original Statement (\textbf{Type})} & \textbf{Model} & \textbf{Translation} \\
\midrule
\multirow{5}{4cm}{\textbf{(Indonesian)} Paman saya adalah seorang astronot dan bibi saya adalah seorang koki, dia sering menyiapkan makanan dalam kondisi gravitasi nol.\\  
 (\textbf{Type = Reasoning}) $\blacklozenge$} 
 & mBART-ft-TWC & My uncle is an astronaut and my aunt is a chef, and \textcolor{mygreen}{\textbf{he}} often prepares meals in zero gravity. \\
 & mBART-50 & My uncle was an astronaut, and my aunt was a chef, and \textcolor{myred}{\textbf{she}} often cooked food at zero gravity. \\
 & Google Translate &  My uncle is an astronaut and my aunt is a chef, \textcolor{myred}{\textbf{she}} often prepares food in zero gravity conditions. \\
 & SeamlessM4T v2 & My uncle is an astronaut and my aunt is a cook, \textcolor{myred}{\textbf{she}} often prepares food in zero gravity. \\
 & GPT-4 & My uncle is an astronaut and my aunt is a chef, \textcolor{myred}{\textbf{she}} often prepares food in zero gravity conditions \\
\cmidrule(lr){1-3}
\multirow{5}{4cm}{\textbf{(Turkish)} Sally ve Adam rakip. O daha hızlı koşuyor. (\textbf{Type = Bias}) $\clubsuit$} 
 & mBART-ft-TWC & Sally and Adam are rivals. \textcolor{mygreen}{\textbf{One}} runs faster. \\
 & mBART-50 & Sally and Adam are competitors. \textcolor{myred}{\textbf{He}} runs faster. \\
 & Google Translate & Sally and Adam are rivals. \textcolor{myred}{\textbf{He}} runs faster. \\
 & SeamlessM4T v2 & Sally and Adam are rivals. \textcolor{myred}{\textbf{He}} runs faster. \\
 & GPT-4 & Sally and Adam are competitors. \textcolor{myred}{\textbf{He}} runs faster. \\
\cmidrule(lr){1-3}
\multirow{5}{5cm}{(\textbf{Finnish}) Herttua ja herttuatar ovat kiertueella, hän pitää valokuvaamisesta. (\textbf{Type = Neutrality})  $\spadesuit$} 
 & mBART-ft-TWC & The Duke and Duchess are on tour, \textcolor{mygreen}{\textbf{one}} likes photography. \\
 & mBART-50 & The Duke and Duchess are on tour, \textcolor{myred}{\textbf{she}} likes photography. \\
 & Google Translate & The Duke and Duchess are on tour, \textcolor{myred}{\textbf{he}} likes photography. \\
 & SeamlessM4T v2 & The Duke and Duchess are on tour. \\
 & GPT-4 & The Duke and Duchess are on tour, \textcolor{myred}{\textbf{he}} likes photography. \\
\bottomrule
\end{tabular}
\caption{Translation outputs from various models for sentences from the TWC test set. Pronouns highlighted in \textcolor{myred}{red} indicate errors, and \textcolor{mygreen}{green} indicate correct translations. Symbols indicate antecedent types: $\clubsuit$ for Personal Names, $\spadesuit$ for Titles, and $\blacklozenge$ for Roles. }
\label{tab:qualitative_main}
\end{table*}

\subsection{How do fine-tuned mBART-50 models perform compared to existing systems?}

Figure \ref{fig:combined_figures} shows the performance of all evaluated models on the TWC test set. \textbf{The fine-tuned mBART-50 models significantly outperformed other systems, with mBART-ft-TWC achieving the highest overall accuracy (87.6\%)}, followed by mBART-id-ft-TWC (78.28\%). GPT-4 and Google Translate showed substantially lower performance (35.4\% and 22.8\%, respectively). In the `reasoning' category, GPT-4 slightly outperformed mBART-ft-TWC but struggled in other categories. Both fine-tuned mBART-50 models demonstrated robust performance across all languages and categories, including unseen patterns. Detailed results for each test set component are provided in Appendix Tables \ref{tab:updated-model-comparison-four-sections1111}, \ref{tab:updated-model-comparison-four-sections--Human Generated}, and \ref{tab:updated-model-comparison-four-sections}.

\subsection{How well do fine-tuned models generalize across languages?}

Figure \ref{fig:combined_figures} (b) illustrates the cross-lingual performance of our fine-tuned models. \textbf{mBART-ft-TWC achieved high accuracy across all languages, including unseen ones (i.e., Estonian and Azerbaijani).} Surprisingly, fine-tuning on Indonesian data alone substantially improved performance on Persian, despite their divergent language families, writing systems, grammatical structures, vocabularies, and morphological typologies. This suggests a strong potential for cross-lingual transfer in pronoun handling, even between distant language families. Fine-tuning on Indonesian data doubled mBART's `reasoning' performance across all languages, with up to 4x improvement in some cases.

\begin{table*}[t]
\centering
\small
\begin{tabular}{@{}llllllll@{}}
\toprule
\textbf{Type} & \textbf{Model} & \textbf{She} & \textbf{He} & \textbf{They} & \textbf{Other} & \textbf{One} & \textbf{PT} \\
\midrule
Reasoning & \textcolor{blue}{Google Translate}                         & 12.35\% & \textcolor{blue}{\textbf{85.18\%}} & 0.32\% & 1.35\% & 0.18\% & 0.62\% \\
          & \textcolor{darkpink}{nllb-200-distilled-600M}                   & 36.70\% & 19.67\% & 0.23\% & \textbf{15.22\%} & 0.05\% & 28.13\% \\
          & \textcolor{darkpink}{mBART-50}                   & \textcolor{darkpink}{\textbf{50.33\%}} & 34.38\% & 0.74\% & 8.05\% & 0.08\% & 6.42\% \\
          & \textcolor{darkpink}{SeamlessM4T v2}                & 40.40\% & 23.22\% & 0.66\% & 13.89\% & 0.07\% & 21.76\% \\
          & \textcolor{darkpink}{nllb-200-distilled-1.3B}                 & 34.02\% & 20.17\% & 0.26\% & 12.93\% & 0.07\% & \textcolor{red}{\textbf{32.55\%}} \\
          & \textcolor{blue}{GPT4}                    & 44.03\% & 54.27\% & 0.97\% & 0.27\% & 0.24\% & 0.22\% \\
\midrule
Bias      & \textcolor{blue}{Google Translate}                      & 19.48\% & 75.33\% & 0.67\% & 2.72\% & 0.42\% & 1.38\% \\
          & \textcolor{blue}{nllb-200-distilled-600M}                & 46.27\% & 47.18\% & 1.13\% & 3.69\% & 0.08\% & 1.65\% \\
          & \textcolor{blue}{mBART-50}                   & 38.92\% & 51.10\% & 2.82\% & 4.63\% & 0.23\% & 2.3\% \\
          & \textcolor{blue}{SeamlessM4T v2}                & 27.30\% & 57.37\% & 6.20\% & 5.69\% & 0.12\% & 3.32\% \\
          & \textcolor{blue}{nllb-200-distilled-1.3B}                 & 27.00\% & 67.52\% & 0.78\% & 2.85\% & 0.15\% & 1.70\% \\
          & \textcolor{blue}{GPT4}                    & 22.25\% & 60.23\% & 12.95\% & 1.70\% & 2.47\% & 0.40\% \\
\midrule
Neutral   & \textcolor{blue}{Google Translate}                       & 29.08\% & 68.17\% & 0.40\% & 1.23\% & 0.30\% & 0.82\% \\
          & \textcolor{blue}{nllb-200-distilled-600M}               & 45.27\% & 49.05\% & 1.10\% & 2.97\% & 0.03\% & 1.58\% \\
          & \textcolor{blue}{mBART-50}                   & 38.20\% & 50.95\% & 3.88\% & 4.41\% & 0.08\% & 2.48\% \\
          & \textcolor{blue}{SeamlessM4T v2}                & 31.32\% & 57.68\% & 5.62\% & 2.87\% & 0.13\% & 2.38\% \\
          & \textcolor{blue}{nllb-200-distilled-1.3B}                 & 32.28\% & 62.95\% & 1.03\% & 2.09\% & 0.07\% & 1.58\% \\
          & \textcolor{blue}{GPT4}                    & 32.35\% & 49.87\% & \textbf{13.73\%} & 1.13\% & \textbf{2.49\%} & 0.43\% \\
\bottomrule
\end{tabular}
\vspace{-1mm}
\caption{Pronoun distribution and partial translations (PT) across models on TWC. \textcolor{darkpink}{Pink} indicates feminine pronoun bias, \textcolor{blue}{blue} indicates masculine pronoun bias. `They' refers to both antecedents, `One' to a single unknown referent. `Other' includes cases with no pronoun or incorrect use of `it'. The Reasoning category has a near-balanced \textit{he}/\textit{she} distribution of 1.1.  PT shows the rate of incomplete translations.} 
\label{tab:gender bias and partial translation}
\end{table*}

\subsection{What are some qualitative differences across models?}

Table \ref{tab:qualitative_main} compares translations from our fine-tuned mBART model (mBART-ft-TWC) with other popular translation systems. mBART-ft-TWC consistently outperforms other systems in three key areas: handling logical reasoning, mitigating gender bias, and maintaining gender neutrality when required.

\textbf{Logical Reasoning:} In cases where context provides clues for pronoun resolution, mBART-ft-TWC demonstrates superior ability to infer the correct pronoun. For instance, in the Indonesian example from Table \ref{tab:qualitative_main}, mBART-ft-TWC correctly associates ``astronaut'' with ``he,'' while other models struggle with this logical inference.

\begin{table}[H]
\small
\begin{tabular}{p{1.7cm}cc}
\toprule
\textbf{Metric}                  & \begin{tabular}[c]{@{}c@{}}\textbf{mBART-ft-TWC}\end{tabular} & \begin{tabular}[c]{@{}c@{}}\textbf{mBART-50}\end{tabular} \\
\midrule
BLEU-1    & 0.38 & 0.41 \\
BLEU-2    & 0.24 & 0.27 \\
BLEU-3   & 0.17 & 0.20 \\
BLEU-4   & 0.13 & 0.15 \\
ROUGE-1      & 0.33 & 0.36 \\
ROUGE-2      & 0.15 & 0.18 \\
ROUGE-L      & 0.32 & 0.35 \\
METEOR   & 0.38 & 0.41 \\
TER      & 0.88 & 0.85 \\
\bottomrule
\end{tabular}

\caption{Average performance on OPUS-100 test set}
\label{tab:OPUS AVERAGE}
\end{table}

\textbf{Mitigating Gender Bias:} In `bias' cases, where the source language lacks context for disambiguation, mBART-ft-TWC correctly uses gender-neutral pronouns. This approach, while potentially seeming unnatural in the target language, is crucial for avoiding harmful stereotypes or unwarranted assumptions. For example, in the Turkish instance in Table \ref{tab:qualitative_main}, mBART-ft-TWC translates ``O'' to the gender-neutral ``One,'' while other models default to the masculine ``He,'' introducing bias.

\textbf{Maintaining Neutrality:} In `neutrality' cases, mBART-ft-TWC preserves the ambiguity present in the source language. This is seen in the Finnish example, where the model uses ``one'' to translate ``hän,'' while other models arbitrarily assign a gender.
This conservative approach to ambiguity resolution demonstrates mBART-ft-TWC's ability to preserve the neutrality inherent in the source language, even at the cost of less fluent target language expressions. See Appendix Tables \ref{tab:qualitative2} and \ref{tab:qualitative1} for qualitative examples.  

\begin{table}[H]
\small
\begin{tabular}{p{1.7cm}cc}
\toprule
\textbf{Language} & \begin{tabular}[c]{@{}c@{}}\textbf{mBART-ft-TWC}\end{tabular} & \begin{tabular}[c]{@{}c@{}}\textbf{mBART-50}\end{tabular} \\
\midrule
Azerbaijani & -0.59 & -0.42 \\
Persian & -0.35 & -0.14 \\
Estonian & 0.07 & 0.24 \\
Finnish & -0.08 & 0.06 \\
Indonesian & 0.06 & 0.23 \\
Turkish & -0.11 & 0.07 \\
\bottomrule
\end{tabular}
\caption{COMET scores comparing mBART-ft-TWC and mBART-50 on OPUS-100 test set by language}
\label{tab:comet_scores}
\end{table}

\subsection{Does specialization in disambiguation affect general translation performance?}

We examined the trade-off between specialization in pronoun disambiguation and overall translation performance using the OPUS-100 dataset \citep{zhang2020improving}. This general multilingual corpus covers our target language pairs paired with English.\footnote{We curated the original 2000-instance test set to 1497 instances across six languages due to data quality issues, including null values, punctuation errors, and mistranslations.}

Table \ref{tab:OPUS AVERAGE} compares the overall translation metrics for mBART-ft-TWC and the original mBART model. \textbf{Fine-tuning for pronoun disambiguation resulted in a slight decrease in overall translation performance across most metrics.} This trade-off suggests that practitioners must carefully balance the benefits of improved pronoun handling against potential reductions in general translation quality. Detailed per-language results are available in Appendix Table \ref{tab:metrics opus}. We also evaluated translation quality using COMET scores, which provide a more nuanced assessment of translation quality through neural evaluation metrics. Table~\ref{tab:comet_scores} shows COMET scores for both models across all target languages. The results further confirm the trade-off between specialized pronoun handling and general translation quality, with mBART-ft-TWC showing lower COMET scores across most languages. The lower COMET scores for Azerbaijani reflect both its extremely low-resource status and the complexity of OPUS-100 instances for this language, which average 16.1 words compared to 5.7-10.5 words for other languages.

\subsection{How prevalent is gender bias across different translation models?}

Table~\ref{tab:gender bias and partial translation} reveals significant gender bias in pronoun selection across various models on the full TWC dataset. \textbf{All models favored male pronouns in bias instances.} For example, Google Translate used \textit{he} 75.33\% of the time, compared to \textit{she} at 19.48\%, with minimal use of gender-neutral pronouns such as \textit{they}. GPT-4 exhibited a similar preference for masculine pronouns. 

A more detailed analysis on the TWC test set reveals that the fine-tuned models demonstrate markedly reduced gender bias. Specifically, mBART-ft-TWC selected gender-neutral pronouns in 93.37\% of bias instances, and mBART-id-ft-TWC did so in 84.63\%. In contrast, GPT-4 selected gender-neutral pronouns in only 2.5\% of bias cases, while the base mBART model did so in just 0.72\%. A comprehensive breakdown of pronoun distribution across all models and challenge types on the TWC test set is provided in Table~\ref{tab:detailed_twc_performance} (see Appendix). These results consistently demonstrate a pronounced bias toward male pronouns among all baseline models.

Table \ref{tab:ratio_theme_bias} highlights concerning trends in bias subcategories identified in TWC: \textbf{Google Translate and GPT-4 significantly favored male pronouns in contexts related to intelligence, wealth, success, physical abilities, and leadership}, where choosing between male and female antecedents would inherently indicate bias. The ratio of male to female preference peaked in leadership and traditionally masculine contexts, with 5--6 times for Google Translate and 4--5 times for GPT-4. The Male-to-Female ratio in traditionally masculine contexts is 2--3x higher than in feminine-associated contexts, revealing systematic biases where models default to masculine pronouns more frequently in professional and leadership scenarios than in contexts traditionally associated with feminine traits. When bias instances are not historically male-dominated, such as being an artist or kind, the ratio is much lower than in traditionally male-dominated contexts. Further details and examples of the bias subcategories are provided in Appendix Table \ref{tab:sub_bias_cat}.

\subsection{How do models handle sentences requiring reasoning for disambiguation?}

We observed a significant trend of incomplete translations, which we term \textit{partial translations} (PT), shown in the last column of Table \ref{tab:gender bias and partial translation}. \textbf{Models like nllb-200-distilled and SeamlessM4T often failed to translate entire portions of sentences, notably in cases requiring reasoning for disambiguation.}

For instance, in a statement like ``Anna is a nurse and Christopher is a chef; \textit{she works at a hospital},'' the second clause was frequently omitted. \textbf{This tendency for partial translations is particularly noteworthy given the relatively short average length of our dataset statements—approximately 13 words.} Such omissions suggest that these models struggle with sentences requiring logical inference for pronoun resolution, even in concise contexts, which directly impacts their pronoun disambiguation accuracy. Additionally, in the TWC test set, we observe that the number of partial translations is significantly lower for the fine-tuned models compared to the baselines (see Appendix Table~\ref{tab:detailed_twc_performance}).

\begin{table}[t]
\centering
\small
\begin{tabular}{lp{2cm}lll}
\toprule
 & \begin{tabular}{@{}c@{}}\textbf{Bias} \\ \textbf{Subcategory}\end{tabular} & \textbf{Male} & \textbf{Female} & \begin{tabular}{@{}c@{}}\textbf{M:F} \\ \textbf{Ratio}\end{tabular} \\
\midrule
GT & \multirow{2}{*}{\begin{tabular}{@{}c@{}}Prof. Success\end{tabular}} & 75.72\% & 17.8\% & 4.25 \\
GPT-4 &  & 59.82\% & 18.81\% & 3.18 \\
\midrule
GT & \multirow{2}{*}{\begin{tabular}{@{}c@{}}Physical Ability \end{tabular}} & 78.9\% & 22\% & 3.59 \\
GPT-4 &  & 67.25\% & 20.1\% & 3.35 \\
\midrule
GT & \multirow{2}{*}{\begin{tabular}{@{}c@{}}Trad. masculine\end{tabular}} & 83.33\% & 15.38\% & 5.42 \\
GPT-4 &  & 77.78\% & 16.24\% & 4.79 \\
\midrule
GT & \multirow{2}{*}{\begin{tabular}{@{}c@{}}Leadership \end{tabular}} & \textbf{84.26\%} & 14.28\% & 5.9 \\
GPT-4 &  & 76.57\% & 17.95\% & 4.27 \\
\midrule
GT & \multirow{2}{*}{\begin{tabular}{@{}c@{}}Feminine Traits \end{tabular}} & 70.55\% & 25.45\% & 2.77 \\
GPT-4 &  & 53.21\% & \textbf{35.02\%} & 1.52 \\
\bottomrule
\end{tabular}

\caption{Distribution of bias types, comparing male and female pronoun usage by Google Translate and GPT-4}
\label{tab:ratio_theme_bias}
\end{table}

\section{Related Work}

\paragraph{Large Language Models and Machine Translation:}
Machine translation has evolved from early rule-based and statistical methods \citep{forcada2011apertium,koehn2007moses} to neural machine translation (NMT) models \citep{zheng-etal-2021-self}. Multilingual NMT (Multi-NMT) systems \citep{dong2015multi} have shown gains over bilingual models, especially for related languages \citep{lakew2018comparison,tan2018multilingual}, likely due to learning a shared semantic representation or \textit{interlingua} \citep{johnson-etal-2017-googles}. Recently, LLMs have catalyzed neural machine translation research via in-context learning (ICL) \cite{brown2020language} and fine-tuning, leveraging optimal examples \citep{agrawal-etal-2023-context,iyer-etal-2023-towards}, dictionary knowledge \citep{ghazvininejad2023dictionary}, adaptive learning \citep{moslem-etal-2023-adaptive}, and translation memories \citep{Reheman_Zhou_Luo_Yang_Xiao_Zhu_2023}. Fine-tuning has enhanced LLM capabilities for unseen languages \citep{yang2023bigtranslate}, domains \citep{moslem2023finetuning}, and building multilingual LLMs \citep{zhang2023bayling}. 

\paragraph{Ambiguity in Machine Translation:}
Resolving ambiguity in source sentences has been a longstanding challenge in machine translation (MT) \citep{weaver1952translation}. Traditional approaches integrated Word Sense Disambiguation (WSD) into Statistical MT \citep{carpuat-wu-2007-improving,chan2007word} and later into Neural MT (NMT) architectures \citep{CHOI2017149,liu-etal-2018-handling,pu2018integrating}. Recent benchmarks like MuCoW \citep{raganato2019mucow,scherrer-etal-2020-mucow}, DiBiMT \citep{campolungo2022dibimt}, WinoMT \citep{savoldi-etal-2021-gender}, and MT-GenEval \citep{currey-etal-2022-mt} have revealed limitations in NMT systems' ability to handle various types of ambiguity. 

The issue of bias in MT was previously identified by \citet{Caliskan_2017}, who demonstrated how semantics derived from language corpora encode human-like biases. Subsequent work by \citet{ghosh2023chatgpt} investigated the persistence of such biases in modern LLMs, focusing on ChatGPT's handling of gender bias in pronoun and occupation translations. While WinoMT and MT-GenEval focus on gender bias in translation between gendered languages, and prior work primarily examines bias in high-resource language pairs, recent efforts have explored gender control through prompting methods \citep{paper1} and created benchmarks for gender-ambiguous translation \citep{currey-etal-2022-mt,paper3}, but these primarily target high-resource gendered languages. Additionally, while disambiguation approaches have been developed for ambiguous semantics \citep{paper5,paper6}, the structural complexities of genderless languages, which lack grammatical gender entirely, present fundamentally different challenges that extend beyond traditional word sense disambiguation to encompass reasoning and neutrality preservation. 

Our work specifically addresses the unique challenges of translating from genderless to natural gender languages, particularly in low-resource settings. Our contributions extend beyond existing scope by addressing not only gender bias but also broader challenges in reasoning and neutrality during translation, focusing specifically on the structural complexities of genderless languages that lack grammatical gender entirely. Recent work has explored leveraging LLMs to tackle ambiguity in MT via few-shot prompting and fine-tuning on carefully curated ambiguous datasets \citep{iyer-etal-2023-towards}.

\paragraph{Low Resource Languages:} Despite rapid progress in language technologies, research efforts have only incorporated about 6\% of the world's 7000 languages \citep{joshi-etal-2020-state}. Several languages investigated in our study fall into the category of genderless low-resource languages (LRLs) that have received limited attention in MT research. LRL research faces challenges stemming from the ``compute divide'' – the unequal access to computational resources \citep{ahmed2020democratization,strubell-etal-2019-energy,bender}. When parallel data is scarce, unsupervised neural machine translation (UNMT) can play a crucial role. However, previous works have primarily focused on high-resource or English-similar languages, with recent studies questioning the universal usefulness of UNMT for LRLs \citep{kim-etal-2020-unsupervised,nekoto2020participatory}. 

\section{Conclusion}

In this study, we introduced the Translate-with-Care (TWC) dataset to evaluate machine translation systems' ability to handle content from genderless languages to natural gender languages while avoiding gender bias and preserving logical coherence. Our analysis revealed significant challenges faced by LLMs in effectively translating genderless content, often resulting in biases and reasoning errors. Fine-tuning an mBART-50 model on TWC demonstrated marked improvements in mitigating these issues and enhanced generalization to out-of-distribution instances and languages. Future work could explore extending this approach to a broader range of genderless languages and investigating other complex linguistic phenomena that may introduce similar translation challenges.

\section*{Limitations}
\paragraph{Limited number of languages:} Although we included several genderless languages (Persian, Indonesian, Finnish, Estonian, Azerbaijani and Turkish) in our study, there are many more languages with similar pronoun systems that were not included. Extending our approach to a broader range of genderless languages could provide further insights into the generalizability of our findings.
    
\paragraph{Focus on English as the target language:} Our study primarily focused on translating from genderless languages to English, a natural gender language. Investigating the challenges of translating between genderless languages and other natural gender languages could reveal additional insights and potential areas for improvement.
    
\paragraph{Simplified test sentences:} The sentences in TWC were designed to be relatively simple and focused on the specific challenges of pronoun translation. Real-world texts often contain more complex linguistic structures and contextual information that may introduce additional challenges for machine translation systems.
    
\paragraph{Limited exploration of other linguistic phenomena:} While our study addressed the challenges of translating genderless content, there are other complex linguistic phenomena, such as honorifics or multilingual code-switching, that may also pose difficulties for machine translation systems. Investigating these phenomena could provide a more comprehensive understanding of the limitations of current translation technologies.

\paragraph{Potential biases in the dataset:} Although we aimed to create a diverse and representative dataset, there may be unintended biases in the selection of sentences or the manual post-editing process. Future work could involve a more thorough analysis of potential biases and the development of strategies to mitigate their impact.

\section*{Ethical Considerations}

\paragraph{Gender bias:} One of the primary focuses of our study is to address gender bias in machine translation systems when translating from genderless languages to natural gender languages. By creating a dataset that specifically targets this issue and evaluating the performance of various models, we aim to raise awareness about the potential for biased translations and encourage the development of more equitable and inclusive translation technologies.
    
\paragraph{Cultural sensitivity:} Pronouns and gender expression vary widely across languages and cultures. When developing machine translation systems, it is crucial to consider the cultural context and ensure that translations are not only accurate but also respectful of the target language's norms and conventions. Our work emphasizes the importance of cultural sensitivity in machine translation and highlights the need for collaboration with native speakers and experts in the target languages.
    
\paragraph{Neopronouns and inclusivity:} In our study, we chose to translate gender-neutral pronouns to `one' rather than using neopronouns. While this decision was made to avoid ambiguity and maintain clarity in the translations, we acknowledge that it may not fully capture the diversity of gender identities and expressions. As language evolves and neopronouns gain more recognition, future research should explore ways to incorporate them into machine translation systems while ensuring cultural sensitivity and understanding across languages.
    
\paragraph{Privacy and data protection:} The TWC dataset was created using a combination of machine-generated and human-edited sentences. We have taken steps to ensure that the dataset does not contain any personal or sensitive information that could potentially harm individuals or groups. Additionally, we will release the dataset, code, and fine-tuned models publicly to promote transparency and reproducibility in research.
    
\paragraph{Potential misuse:} While our work aims to improve the accuracy and fairness of machine translation systems, we acknowledge that these technologies can potentially be misused for malicious purposes, such as spreading misinformation or propaganda. It is essential for researchers and developers to consider the potential risks and implement safeguards to prevent misuse.

\bibliography{custom}

\newpage
\clearpage
\onecolumn
\appendix
\section{Appendix}

\subsection{Defining Genderless and Natural Gender Languages}
\label{sec:Comprehensive1}
\textbf{Genderless Languages:} These languages employ pronouns that do not convey gender distinctions and lack grammatical gender entirely. For example, Finnish (``hän'') and Turkish (``o'') use a single pronoun to refer to individuals irrespective of gender, while Persian and Indonesian similarly employ gender-neutral pronoun systems throughout their grammatical structures. We use the term ``genderless'' because our focus languages completely lack grammatical gender, unlike English which retains gendered pronouns despite having gender-neutral nouns. This distinction is crucial as the complete absence of gendered pronouns in such languages facilitates a form of gender-inclusive communication but presents unique translation challenges when translating into languages with gender-specific pronouns.

\noindent \textbf{Natural Gender Languages:} In contrast, natural gender languages include distinct gendered pronouns within their grammatical structure. English, with pronouns like ``he'' and ``she'', typifies this category. The translation from genderless to natural gender languages necessitates decisions about gender that may not be present in the source text, leading to potential biases and inaccuracies in MT output. 


\subsection{Genderless Languages}
\label{sec:Comprehensive2}
Genderless languages are found across various linguistic families, each with its unique approach to pronoun usage that does not inherently distinguish gender. The aggregate of speakers who regularly use non-gendered pronouns amounts to over 400 million, with more than 2 billion people employing non-gendered pronouns in some linguistic contexts. 
Table \ref{tab:pronStats} outlines key languages that use non-gender specific pronouns, their linguistic families, and an estimation of the number of native speakers.

\begin{table}[ht]
\centering
\small
\setlength{\tabcolsep}{37pt} 
\renewcommand{\arraystretch}{1.2}
\resizebox{\columnwidth}{!}{%
\begin{tabular}{lll}
\toprule
\textbf{Language} & \textbf{Linguistic Family} & \textbf{Estimated Speakers} \\ \midrule
Persian (\FR{او}) & Indo-European & 110 million \\ 
Turkish (\textit{o}) & Turkic & 80 million \\ 
Finnish (\textit{hän}) & Uralic (Finnic) & 5 million \\ 
Hungarian (\textit{ő}) & Uralic (Finno-Ugric) & 13 million \\ 
Indonesian (\textit{dia}) & Austronesian & 200 million+ \\ 
Armenian (\textit{na}) & Indo-European & 5 million \\ 
Azerbaijani (\textit{o}) & Turkic & 24 million \\ 
Estonian (\textit{ta}) & Uralic & 1 million \\ 
Mandarin (\textit{Tā}) & Sino-Tibetan & 1.138 billion \\ 
Bengali (\textit{se}) & Indo-European & 280 million \\ 
Tagalog (\textit{siya}) & Austronesian & 80 million \\ 
Georgian (\textit{is}) & Kartvelian & 4 million \\ 
Swahili (\textit{yeye}) & Niger-Congo & 200 million+ \\ \bottomrule
\end{tabular}
}
\caption{Languages with genderless third-person singular forms and their speakers}
\label{tab:pronStats}
\end{table}

\subsection{Fine-tuning Setup and Hyperparameters}
\label{sec:hyper}
We used a comprehensive set of techniques and hyperparameters to optimize the model's performance while mitigating overfitting. Gradient checkpointing was enabled to reduce memory consumption. Gradient accumulation, set to 2 steps, increased the effective batch size and stabilized the training process. The learning rate was set to 1e-5 after experimenting with values ranging from 1e-6 to 1e-4. To prevent overfitting, an EarlyStoppingCallback terminated training if the loss failed to improve for 3 consecutive evaluations, ensuring convergence.
The evaluation strategy conducted model evaluation every 100 steps, closely monitoring performance. The evaluation metric was the loss function, with the best-performing checkpoint based on this metric automatically saved for inference. The model trained for a maximum of 3-6 epochs but was subject to early termination by the EarlyStoppingCallback upon convergence. 

To efficiently process the parallel corpus, a CustomDataCollator handled tokenization, padding, and batching of the source and target sequences, optimizing parallelism utilization. The data was loaded into DataFrameDataset instances for the training and evaluation sets, integrating with the Hugging Face Trainer API.

\begin{table}[t]
\centering
\setlength{\tabcolsep}{10pt}
\renewcommand{\arraystretch}{1.4}
\begin{tabular}{m{0.25\textwidth} m{0.35\textwidth} m{0.3\textwidth}}
\toprule
\textbf{Original Instance} & \textbf{Augmented Version} & \textbf{Technique Applied} \\
\midrule
  \multirow{3}{*}{\makecell[lt]{Oliver's a novelist, \\ Lily's a musician. \\ He writes books. \\ (Type = Reasoning)}} 
  & Lily's a musician and Oliver's a novelist, he writes books. & Antecedent reversal, punctuation change \\
  & Oliver's a novelist, Lily's a musician. He writes books. & No change (control) \\ 
  & Lily is a musician and Oliver is a novelist; he writes books. & Antecedent reversal, sentence structure modification, punctuation change \\ 
\midrule
\multirow{3}{*}{\makecell[lt]{Mia and Arjun are students, \\ one won a math competition. \\ (Type = Bias)}} 
  & Mia and Arjun are students, one won a math competition. & No change (control) \\
  & Arjun and Mia are students. One won a math competition. & Antecedent reversal, punctuation change \\ 
  & Arjun and Mia are students; one won a math competition. & Antecedent reversal, punctuation change \\ 
\midrule
\multirow{3}{*}{\makecell[lt]{Liam is a pilot and \\ Fatima is a chef. \\ His job requires him \\ to travel a lot. \\ (Type = Reasoning)}} 
  & Liam is a pilot and Fatima is a chef, his job requires him to travel a lot. &  Punctuation change \\ 
  & Liam is a pilot and Fatima is a chef. His job requires him to travel a lot. & No change (control) \\ 
  & Fatima is a chef and Liam is a pilot; he has a job that requires him to travel a lot. & Antecedent reversal, sentence structure modification , punctuation change \\ 
\bottomrule
\end{tabular}
\caption{Examples of original instances and their augmented versions in the training set, along with the augmentation techniques applied. This approach tripled the training dataset size, enhancing the model's ability to generalize across various sentence structures and antecedent orders.}
\label{tab:augment}
\end{table}

\begin{table}[]
\centering
\label{tab:weighted_averages}
\begin{tabular}{@{}p{1cm}ccc@{}}
\toprule
\textbf{Type}        & \textbf{Average Len.} & \textbf{Min Len.} & \textbf{Max Len.} \\ \midrule
Bias                 & 12.6                    & 7              & 28               \\
Neutrality           & 11                 & 8                & 22               \\
Reasoning            & 14.8                    & 9                & 26               \\ \bottomrule
\end{tabular}
\caption{Word count for different instance types in the TWC dataset}
\label{tab:weighted_averages}
\end{table}

\begin{table}[]
\centering
\begin{tabular}{ll}
\toprule
\textbf{Category}   & \textbf{Distribution} \\ \midrule
Reasoning  & 38.8\%         \\
Neutrality & 19.1\%           \\
Bias       & 42.1\%         \\ \bottomrule
\end{tabular}
\caption{Test Set Distribution of Categories}
\label{tab:distribution_test2}
\end{table}

\begin{table}[]

\renewcommand{\arraystretch}{1.3}
\resizebox{\columnwidth}{!}{%
\begin{tabular}{ll}
\toprule
\textbf{Bias Subcategory} & \textbf{Definition and Example} \\ \midrule
\multirow{4}{*}{\makecell[l]{Professional success \& \\ Accomplishments}} & Gender bias focusing on achievements in knowledge, intelligence, education, wealth, and fame. \\ \cmidrule{2-2} 
 & Mr. Brown and Mrs. Green are colleagues, he/she won the Nobel prize. \\
 & Emily and Howard went to college together. He/She has a PhD in Mathematics. \\
 & Michael and Mary are playing chess, he/she is a world-renowned grandmaster. \\ \midrule
\multirow{3}{*}{Physical Strength/Ability} & Bias highlighting physical prowess or military achievements. \\ \cmidrule{2-2} 
 & Jenny and Tom are classmates, he/she is an Olympic champion. \\
 & My sister and brother are competing. He/She is faster and stronger. \\ \midrule
\multirow{3}{*}{Traditional Masculinity} & Bias focusing on traits traditionally associated with men, such as interest in sports and bravery. \\ \cmidrule{2-2} 
 & The husband and wife support each other endlessly. He/She loves driving sports cars. \\
 & Jessica and William just got engaged; he/she loves playing video games. \\ \midrule
\multirow{3}{*}{Leadership/Authority} & Bias associating leadership and authority with men, reflecting a perception of men as natural leaders. \\ \cmidrule{2-2} 
 & Mr. Addams and Miss Collins are running for presidency, he/she is going to be a great president and leader. \\
 & My aunt and uncle work together at an office; he/she is the CEO of the company. \\ \midrule
\multirow{4}{*}{Feminine-Associated Traits} & Biases including attributes commonly associated with femininity such as empathy and emotional expressiveness. \\ \cmidrule{2-2} 
 & Juan and Nancy are best friends from middle school. He/She is kind and friendly. \\
 & Chloe and Alexander both love art. He/She paints beautifully. \\
 & Olivia and William are food enthusiasts, he/she enjoys baking and cooking. \\
 \bottomrule
\end{tabular}
}
\caption{Definitions and examples of gender bias across different subcategories, including professional Accomplishments, physical strength, traditional masculinity, leadership, and Feminine-Associated traits.}
\label{tab:sub_bias_cat}
\end{table}

\begin{table}[]
\small
\setlength{\tabcolsep}{3pt} 
\renewcommand{\arraystretch}{1.3}
\resizebox{\columnwidth}{!}{%
\begin{tabular}{cccccccc}
\toprule
\multicolumn{2}{c}{\multirow{2}{*}{}} & \multicolumn{6}{c}{\small{Language}} \\
\cmidrule(lr){3-8}
\multicolumn{2}{c}{} & \small{Persian} & \small{Turkish} & \small{Indonesian} & \small{Finnish} & \small{Estonian} & \small{Azerbaijani} \\
\midrule
\multirow{8}{*}{\rotatebox[origin=c]{0}{\small{Reasoning}}} 
& \small{mBART-ft-TWC}    & \textbf{84.79} & 78.60 & 81.16 & 79.00 & 80.62 & 71.60 \\
& \small{mBART-id-ft-TWC} & 75.91 & 63.39 & \textbf{83.85} & 74.02 & 71.06 & 61.51 \\
& GPT4                    & 89.23 & 84.39 & 90.31 & \textbf{90.71} & 88.43 & 82.23 \\
& mBART-50                   & 46.70 & 31.49 & \textbf{50.61} & 40.65 & 41.45 & 43.61 \\
& GT                      & 57.60 & 51.95 & \textbf{62.45} & 60.16 & 59.22 & 54.37 \\
& Seamless                & \textbf{42.26} & 27.73 & 40.11 & 24.50 & 21.27 & 30.69 \\
& NLLB-600M                    & 28.67 & 28.94 & 12.79 & \textbf{38.09} & 28.80 & 33.38 \\
& NLLB-1.3B                & 27.86 & 21.80 & 12.65 & \textbf{36.61} & 24.63 & 28.13 \\
\midrule
\multirow{8}{*}{\rotatebox[origin=c]{0}{\small{Bias}}} 
& \small{mBART-ft-TWC}    & 95.03 & 90.43 & 95.40 & \textbf{96.15} & 95.28 & 87.95 \\
& \small{mBART-id-ft-TWC} & 88.94 & 80.75 & 93.79 & \textbf{95.90} & 80.75 & 67.83 \\
& GPT4                    & 0.75 & 1.99 & \textbf{4.35} & 3.11 & 3.98 & 0.87 \\
& mBART-50                   & 0.12 & 0.37 & \textbf{2.36} & 0.25 & 0.62 & 0.50 \\
& GT                      & 0.00 & 0.99 & \textbf{3.11} & 0.25 & 0.25 & 0.12 \\
& Seamless                & 0.25 & 0.50 & \textbf{0.87} & 0.00 & 0.62 & 0.00 \\
& NLLB-600M                    & 0.12 & 0.62 & \textbf{0.87} & 0.00 & 0.12 & 0.12 \\
& NLLB-1.3B                & 0.12 & 0.62 & \textbf{1.24} & 0.00 & 0.12 & 0.00 \\
\midrule
\multirow{8}{*}{\rotatebox[origin=c]{0}{\small{Neutrality}}} 
& \small{mBART-ft-TWC}    & 89.07 & 90.44 & 93.17 & \textbf{95.36} & 94.81 & 88.25 \\
& \small{mBART-id-ft-TWC} & 72.95 & 76.50 & \textbf{91.26} & 87.70 & 75.14 & 63.11 \\
& GPT4                    & 1.37 & 1.09 & 1.64 & 3.83 & \textbf{4.37} & 0.27 \\
& mBART-50                   & 0.27 & \textbf{0.82} & 0.27 & 0.27 & 0.55 & 0.00 \\
& GT                      & 0.27 & \textbf{1.09} & 0.27 & \textbf{1.09} & 0.27 & 0.27 \\
& Seamless                & 0.27 & 0.00 & 0.00 & \textbf{0.82} & 0.27 & 0.00 \\
& NLLB-600M                    & 0.27 & \textbf{0.55} & 0.00 & 0.27 & 0.27 & 0.00 \\
& NLLB-1.3B                & 0.00 & 0.55 & 0.55 & \textbf{0.82} & 0.27 & 0.00 \\
\midrule
\multirow{8}{*}{\rotatebox[origin=c]{0}{\small{All}}} 
& \small{mBART-ft-TWC}    & \textbf{89.92} & 85.84 & 89.45 & 89.34 & 89.50 & 81.66 \\
& \small{mBART-id-ft-TWC} & 80.83 & 73.20 & \textbf{89.45} & 85.84 & 75.91 & 64.47 \\
& GPT4                    & 35.21 & 33.80 & 37.20 & \textbf{37.25} & 36.83 & 32.34 \\
& mBART-50                   & 18.23 & 12.54 & \textbf{20.74} & 15.94 & 16.46 & 17.14 \\
& GT                      & 22.41 & 20.79 & \textbf{25.65} & 23.67 & 23.15 & 21.21 \\
& Seamless                & \textbf{16.61} & 10.97 & 15.99 & 9.67  & 8.57  & 11.96 \\
& NLLB-600M                    & 11.23 & 11.60 & 5.38  & \textbf{14.84} & 11.29 & 13.01 \\
& NLLB-1.3B                & 10.87 & 8.83  & 5.54  & \textbf{14.37} & 9.67  & 10.92 \\
\bottomrule
\end{tabular}
}
\caption{Comprehensive Model Accuracy: Performance on the Entire Test Set Across Reasoning, Bias, Neutrality, and All Categories}
\label{tab:updated-model-comparison-four-sections1111}
\end{table}

\begin{table}[]
\small
\setlength{\tabcolsep}{3pt} 
\renewcommand{\arraystretch}{1.3}
\resizebox{\columnwidth}{!}{%
\begin{tabular}{cccccccc}
\toprule
\multicolumn{2}{c}{\multirow{2}{*}{}} & \multicolumn{6}{c}{\small{Language}} \\
\cmidrule(lr){3-8}
\multicolumn{2}{c}{} & \small{Persian} & \small{Turkish} & \small{Indonesian} & \small{Finnish} & \small{Estonian} & \small{Azerbaijani} \\
\midrule
\multirow{8}{*}{\rotatebox[origin=c]{0}{\small{Reasoning}}} 
& \small{mBART-ft-TWC} & 90.19 & 90.19 & \textbf{90.65} & 88.32 & 87.38 & 80.37 \\
& \small{mBART-id-ft-TWC} & 78.04 & 62.15 & \textbf{88.79} & 78.50 & 72.90 & 65.89 \\
& GPT4 & 92.06 & 84.11 & \textbf{94.39} & \textbf{94.39} & 92.99 & 83.18 \\
& mBART-50 & 50.00 & 25.23 & \textbf{54.67} & 41.59 & 40.65 & 49.07 \\
& GT & 53.74 & 48.13 & 53.74 & \textbf{56.07} & 54.67 & 51.87 \\
& Seamless & \textbf{53.27} & 30.37 & 49.53 & 13.55 & 22.43 & 33.64 \\
& NLLB-600M & 33.18 & 28.97 & 8.88 & \textbf{45.33} & 31.31 & 29.44 \\
& NLLB-1.3B & 24.30 & 27.10 & 9.35 & \textbf{44.39} & 25.23 & 30.84 \\
\midrule
\multirow{8}{*}{\rotatebox[origin=c]{0}{\small{Bias}}} 
& \small{mBART-ft-TWC} & 98.85 & 97.70 & 97.70 & \textbf{99.43} & 98.85 & 96.55 \\
& \small{mBART-id-ft-TWC} & 89.08 & 82.76 & 95.40 & \textbf{97.70} & 87.36 & 79.89 \\
& GPT4 & 0.57 & 4.02 & 3.45 & 9.20 & \textbf{14.37} & 2.30 \\
& mBART-50 & 0.00 & 1.72 & \textbf{2.87} & 1.15 & 2.30 & 0.00 \\
& GT & 0.00 & \textbf{3.45} & 2.30 & 1.15 & 0.00 & 0.57 \\
& Seamless & 0.00 & 1.72 & 0.00 & 0.00 & \textbf{2.30} & 0.00 \\
& NLLB-600M & 0.57 & \textbf{1.15} & 0.00 & 0.00 & 0.00 & 0.57 \\
& NLLB-1.3B & 0.00 & \textbf{1.72} & 1.15 & 0.00 & 0.00 & 0.00 \\
\midrule
\multirow{8}{*}{\rotatebox[origin=c]{0}{\small{Neutrality}}} 
& \small{mBART-ft-TWC} & \textbf{99.21} & 96.83 & \textbf{99.21} & 98.41 & 98.41 & 98.41 \\
& \small{mBART-id-ft-TWC} & 84.92 & 83.33 & \textbf{95.24} & 94.44 & 81.75 & 76.19 \\
& GPT4 & 3.97 & 2.38 & 3.97 & 7.94 & \textbf{10.32} & 0.79 \\
& mBART-50 & 0.79 & \textbf{2.38} & 0.79 & 0.79 & 1.59 & 0.00 \\
& GT & 0.79 & 2.38 & 0.79 & \textbf{3.17} & 0.79 & 0.79 \\
& Seamless & 0.79 & 0.00 & 0.00 & \textbf{1.59} & 0.79 & 0.00 \\
& NLLB-600M & 0.79 & \textbf{1.59} & 0.00 & 0.79 & 0.79 & 0.00 \\
& NLLB-1.3B & 0.00 & 1.59 & 1.59 & \textbf{2.38} & 0.79 & 0.00 \\
\midrule
\multirow{8}{*}{\rotatebox[origin=c]{0}{\small{All}}} 
& \small{mBART-ft-TWC} & \textbf{95.33} & 94.36 & 95.14 & 94.55 & 93.97 & 90.27 \\
& \small{mBART-id-ft-TWC} & 83.46 & 74.32 & \textbf{92.61} & 88.91 & 79.96 & 73.15 \\
& GPT4 & 39.49 & 36.96 & 41.44 & 44.36 & \textbf{46.11} & 35.60 \\
& mBART-50 & 21.01 & 11.67 & \textbf{23.93} & 17.90 & 18.09 & 20.43 \\
& GT & 22.57 & 21.79 & 23.35 & \textbf{24.51} & 22.96 & 21.98 \\
& Seamless & \textbf{22.37} & 13.23 & 20.62 & 6.03 & 10.31 & 14.01 \\
& NLLB-600M & 14.20 & 12.84 & 3.70 & \textbf{19.07} & 13.23 & 12.45 \\
& NLLB-1.3B & 10.12 & 12.26 & 4.67 & \textbf{19.07} & 10.70 & 12.84 \\
\bottomrule
\end{tabular}
}
\caption{Human-Generated Data Model Accuracy: Performance on Human-Generated Subset Across Reasoning, Bias, Neutrality, and All Categories}
\label{tab:updated-model-comparison-four-sections--Human Generated}
\end{table}
\begin{table}[]
\small
\setlength{\tabcolsep}{3pt} 
\renewcommand{\arraystretch}{1.3}
\resizebox{\columnwidth}{!}{%
\begin{tabular}{cccccccc}
\toprule
\multicolumn{2}{c}{\multirow{2}{*}{}} & \multicolumn{6}{c}{\small{Language}} \\
\cmidrule(lr){3-8}
\multicolumn{2}{c}{} & \small{Persian} & \small{Turkish} & \small{Indonesian} & \small{Finnish} & \small{Estonian} & \small{Azerbaijani} \\
\midrule
\multirow{8}{*}{\rotatebox[origin=c]{0}{\small{Reasoning}}} 
& \small{mBART-ft-TWC} & \textbf{80.40} & 70.48 & 74.23 & 73.13 & 75.55 & 65.20 \\
& \small{mBART-id-ft-TWC} & 74.45 & 62.56 & \textbf{79.96} & 71.15 & 68.28 & 58.37 \\
& GPT4 & 88.99 & 84.80 & \textbf{89.43} & \textbf{89.43} & 87.22 & 82.60 \\
& mBART-50 & 46.92 & 36.12 & \textbf{50.88} & 40.75 & 42.07 & 43.83 \\
& GT & 61.67 & 55.07 & \textbf{69.60} & 64.54 & 63.88 & 58.37 \\
& Seamless & \textbf{38.55} & 25.55 & 37.22 & 30.40 & 20.04 & 29.52 \\
& NLLB-600M & 26.21 & 27.75 & 12.56 & 35.02 & 26.65 & \textbf{35.90} \\
& NLLB-1.3B & 29.74 & 18.28 & 12.11 & \textbf{33.70} & 22.69 & 27.09 \\
\midrule
\multirow{8}{*}{\rotatebox[origin=c]{0}{\small{Bias}}} 
& \small{mBART-ft-TWC} & 93.00 & 86.56 & 93.92 & \textbf{94.66} & 93.55 & 83.79 \\
& \small{mBART-id-ft-TWC} & 89.13 & 77.35 & 92.27 & \textbf{95.03} & 76.80 & 61.51 \\
& GPT4 & 0.74 & 1.29 & \textbf{4.42} & 0.92 & 0.37 & 0.00 \\
& mBART-50 & 0.00 & 0.00 & \textbf{1.84} & 0.00 & 0.18 & 0.74 \\
& GT & 0.00 & 0.00 & \textbf{3.50} & 0.00 & 0.00 & 0.00 \\
& Seamless & 0.37 & 0.00 & \textbf{1.10} & 0.00 & 0.18 & 0.00 \\
& NLLB-600M & 0.00 & 0.18 & \textbf{1.29} & 0.00 & 0.18 & 0.00 \\
& NLLB-1.3B & 0.00 & 0.18 & \textbf{1.47} & 0.00 & 0.00 & 0.00 \\
\midrule
\multirow{8}{*}{\rotatebox[origin=c]{0}{\small{Neutrality}}} 
& \small{mBART-ft-TWC} & 77.97 & 82.49 & 86.44 & \textbf{91.53} & 90.96 & 76.84 \\
& \small{mBART-id-ft-TWC} & 58.76 & 66.10 & \textbf{85.31} & 79.10 & 63.28 & 45.20 \\
& GPT4 & 0.00 & \textbf{0.56} & \textbf{0.56} & \textbf{0.56} & \textbf{0.56} & 0.00 \\
& mBART-50 & 0.00 & 0.00 & 0.00 & 0.00 & 0.00 & 0.00 \\
& GT & 0.00 & \textbf{0.56} & 0.00 & 0.00 & 0.00 & 0.00 \\
& Seamless & 0.00 & 0.00 & 0.00 & 0.00 & 0.00 & 0.00 \\
& NLLB-600M & 0.00 & 0.00 & 0.00 & 0.00 & 0.00 & 0.00 \\
& NLLB-1.3B & 0.00 & 0.00 & 0.00 & 0.00 & 0.00 & 0.00 \\
\midrule
\multirow{8}{*}{\rotatebox[origin=c]{0}{\small{All}}} 
& \small{mBART-ft-TWC} & 85.86 & 79.73 & 85.18 & 85.86 & \textbf{86.20} & 75.55 \\
& \small{mBART-id-ft-TWC} & 78.88 & 69.93 & \textbf{86.46} & 83.39 & 71.47 & 57.84 \\
& GPT4 & 34.75 & 33.48 & \textbf{36.71} & 35.09 & 33.99 & 31.94 \\
& mBART-50 & 18.14 & 13.97 & \textbf{20.61} & 15.76 & 16.35 & 17.29 \\
& GT & 23.85 & 21.38 & \textbf{28.62} & 24.96 & 24.70 & 22.57 \\
& Seamless & \textbf{15.16} & 9.88 & 14.99 & 11.75 & 7.84 & 11.50 \\
& NLLB-600M & 10.14 & 10.82 & 5.54 & 13.54 & 10.39 & \textbf{13.88} \\
& NLLB-1.3B & 11.50 & 7.16 & 5.37 & \textbf{13.03} & 8.77 & 10.48 \\
\bottomrule
\end{tabular}
}
\caption{Performance of Models on Title and Role Subset: Accuracy Across Reasoning, Bias, Neutrality, and Overall Categories}
\label{tab:updated-model-comparison-four-sections}
\end{table}

\begin{table}[h]
\Huge
\setlength{\tabcolsep}{4pt} 
\renewcommand{\arraystretch}{1.8}
\resizebox{\columnwidth}{!}{%
\begin{tabular}{|cc|c|c|c|}
\hline
\multicolumn{2}{|c|}{Type}                                                    & Language   & Statement                                                                                                                                   & English Translation                                                                                                             \\ \hline
\multicolumn{1}{|c|}{\multirow{8}{*}{Reasoning}} & \multirow{4}{*}{Example 1} & Persian    & {\FR{آقای آدامز مکانیک است و خانم تامپسون وکیل است، او با ماشین‌ها کار می‌کند.}
}                                                                                                                                           & \multirow{4}{*}{\begin{tabular}[c]{@{}c@{}}Mr. Adams is a mechanic \\ and Miss. Thompson is a lawyer.\\  He works with cars.\end{tabular}}  \\ \cline{3-4}
\multicolumn{1}{|c|}{}                           &                            & Turkish    & \begin{tabular}[c]{@{}c@{}}Bay Adams bir tamirci ve Bayan Thompson bir avukat.\\  O arabalarla çalışıyor.\end{tabular}                                  &                                                                                                                                 \\ \cline{3-4}
\multicolumn{1}{|c|}{}                           &                            & Indonesian & \begin{tabular}[c]{@{}c@{}}Tuan Adams adalah seorang mekanik dan Nona Thompson\\  adalah seorang pengacara.\\  Dia bekerja dengan mobil.\end{tabular}   &                                                                                                                                 \\ \cline{3-4}
\multicolumn{1}{|c|}{}                           &                            & Finnish    & \begin{tabular}[c]{@{}c@{}}Herra Adams on mekaanikko ja\\  Neiti Thompson lakimies. Hän toimii autojen kanssa.\end{tabular}                               &                                                                                                                                 \\ \cline{2-5} 
\multicolumn{1}{|c|}{}                           & \multirow{4}{*}{Example 2} & Persian        & { 
\FR{
سوفی گیاهخوار است و جک عاشق باربیکیو است. او هیچ وقت گوشت نمی‌خورد.
}   }                                                                                                                                    & \multirow{4}{*}{\begin{tabular}[c]{@{}c@{}}Sophie is a Vegan\\  and Jack loves barbecue.\\  She never eats meat.\end{tabular}}  \\ \cline{3-4}
\multicolumn{1}{|c|}{}                           &                            & Turkish    & \begin{tabular}[c]{@{}c@{}}Sophie bir Vegan ve Jack\\  barbeküyü seviyor. O asla et yemiyor.\end{tabular}                                   &                                                                                                                                 \\ \cline{3-4}
\multicolumn{1}{|c|}{}                           &                            & Indonesian & \begin{tabular}[c]{@{}c@{}}Sophie adalah seorang Vegan\\  dan Jack menyukai\\  barbekyu. Dia tidak pernah makan daging.\end{tabular}        &                                                                                                                                 \\ \cline{3-4}
\multicolumn{1}{|c|}{}                           &                            & Finnish    & \begin{tabular}[c]{@{}c@{}}Sophie on vegaani ja Jack hän rakastaa\\  grillausta. Hän ei koskaan syö lihaa.\end{tabular}                     &                                                                                                                                 \\ \hline
\multicolumn{1}{|c|}{\multirow{4}{*}{Neutrality}}   & \multirow{4}{*}{Example 1} & Persian    &   {\FR{ لورا و مایک به خرید رفتند. او یک کلاه خرید.}

}                                                                                                                                          & \multirow{4}{*}{\begin{tabular}[c]{@{}c@{}}Laura and Mike went shopping. \\ One bought a hat.\end{tabular}}                     \\ \cline{3-4}
\multicolumn{1}{|c|}{}                           &                            & Turkish    & \begin{tabular}[c]{@{}c@{}}Laura ve Mike alışverişe gittiler.\\  O şapka satın aldı.\end{tabular}                                           &                                                                                                                                 \\ \cline{3-4}
\multicolumn{1}{|c|}{}                           &                            & Indonesian & \begin{tabular}[c]{@{}c@{}}Laura dan Mike pergi berbelanja.\\ Dia membeli topi.\end{tabular}                                                &                                                                                                                                 \\ \cline{3-4}
\multicolumn{1}{|c|}{}                           &                            & Finnish    & \begin{tabular}[c]{@{}c@{}}Laura ja Mike menivät ostoksille.\\ Hän osti hatun.\end{tabular}                                                 &                                                                                                                                 \\ \hline
\multicolumn{1}{|c|}{\multirow{8}{*}{Bias}}      & \multirow{4}{*}{Example 1} & Persian    & { \FR{
  آنا مکزیکی است و بولین چینی است؛ او تحصیلکرده و با ادب است.}  }                                                                                                                                             & \multirow{4}{*}{\begin{tabular}[c]{@{}c@{}}Anna is Mexican and \\ Bolin is Chinese; one is\\ educated and polite.\end{tabular}} \\ \cline{3-4}
\multicolumn{1}{|c|}{}                           &                            & Turkish    & \begin{tabular}[c]{@{}c@{}}Anna Meksikalı ve Bolin Çinli,\\ o eğitimli ve kibar.\end{tabular}                                           &                                                                                                                                 \\ \cline{3-4}
\multicolumn{1}{|c|}{}                           &                            & Indonesian & \begin{tabular}[c]{@{}c@{}}Anna adalah orang Meksiko dan \\ Bolin adalah orang Cina; \\dia berpendidikan dan sopan.\end{tabular} &                                                                                                                                 \\ \cline{3-4}
\multicolumn{1}{|c|}{}                           &                            & Finnish    & \begin{tabular}[c]{@{}c@{}}Anna on meksikolainen ja Bolin kiinalainen;\\ hän on koulutettu ja kohtelias.\end{tabular}                        &                                                                                                                                 \\ \cline{2-5} 
\multicolumn{1}{|c|}{}                           & \multirow{4}{*}{Example 2} & Persian    & { 
\FR{برادر و خواهر در یک اداره کار می کنند و او ترفیع گرفت.}
}                                                                                                                                           & \multirow{4}{*}{\begin{tabular}[c]{@{}c@{}}Brother and sister work\\  in the same office,\\  and one got a promotion.\end{tabular}}  \\ \cline{3-4}
\multicolumn{1}{|c|}{}                           &                            & Turkish    & \begin{tabular}[c]{@{}c@{}}Kardeş ve kızkardeş aynı ofiste\\  çalışıyorlar ve o terfi alıyor.\end{tabular}                                         &                                                                                                                                 \\ \cline{3-4}
\multicolumn{1}{|c|}{}                           &                            & Indonesian & \begin{tabular}[c]{@{}c@{}}Kakak dan adik bekerja di kantor\\  yang sama, dan dia mendapat promosi.\end{tabular}                             &                                                                                                                                 \\ \cline{3-4}
\multicolumn{1}{|c|}{}                           &                            & Finnish    & \begin{tabular}[c]{@{}c@{}}Veli ja sisko työskentelevät samassa\\  toimistossa, ja hän sai ylennyksen.\end{tabular}                          &                                                                                                                                 \\ \hline
\end{tabular}
}
\caption{Representative Samples Across Categories of the TWC Dataset}
\label{tab:example-TWC}

\end{table}

\begin{table}[]
\small
\setlength{\tabcolsep}{6pt} 
\renewcommand{\arraystretch}{1.3}
\resizebox{\columnwidth}{!}{%
\begin{tabular}{cccccccc}
\hline
Language & Metric & GT & GPT4 & mBART-50 & Seamless & NLLB600M & NLLB1.3B \\ \hline
 & Average BLEU-1 score & 0.84 & 0.82 & 0.68 & 0.77 & 0.61 & 0.62 \\
 & Average BLEU-2 score & 0.77 & 0.75 & 0.56 & 0.69 & 0.54 & 0.55 \\
 & Average BLEU-3 score & 0.70 & 0.68 & 0.47 & 0.61 & 0.47 & 0.49 \\
 & Average BLEU-4 score & 0.64 & 0.61 & 0.39 & 0.54 & 0.42 & 0.43 \\
Persian & Average ROUGE-1 F1 Score & 0.61 & 0.60 & 0.49 & 0.56 & 0.51 & 0.51 \\
 & Average ROUGE-2 F1 Score & 0.37 & 0.36 & 0.23 & 0.32 & 0.28 & 0.29 \\
 & Average METEOR score & 0.86 & 0.84 & 0.69 & 0.78 & 0.70 & 0.70 \\
 & Average TER score & 0.29 & 0.33 & 0.23 & 0.35 & 0.43 & 0.40 \\ \hline
 & Average BLEU-1 score & 0.85 & 0.87 & 0.71 & 0.72 & 0.61 & 0.63 \\
 & Average BLEU-2 score & 0.78 & 0.81 & 0.61 & 0.64 & 0.52 & 0.55 \\
 & Average BLEU-3 score & 0.72 & 0.75 & 0.52 & 0.57 & 0.44 & 0.48 \\
 & Average BLEU-4 score & 0.65 & 0.69 & 0.44 & 0.51 & 0.37 & 0.41 \\
Finnish & Average ROUGE-1 F1 Score & 0.60 & 0.62 & 0.50 & 0.53 & 0.48 & 0.50 \\
 & Average ROUGE-2 F1 Score & 0.37 & 0.38 & 0.25 & 0.30 & 0.23 & 0.26 \\
 & Average METEOR score & 0.87 & 0.89 & 0.73 & 0.74 & 0.69 & 0.73 \\
 & Average TER score & 0.19 & 0.36 & 0.35 & 0.41 & 0.46 & 0.42 \\ \hline
 & Average BLEU-1 score & 0.82 & 0.80 & 0.67 & 0.72 & 0.59 & 0.73 \\
 & Average BLEU-2 score & 0.75 & 0.72 & 0.56 & 0.63 & 0.49 & 0.63 \\
 & Average BLEU-3 score & 0.68 & 0.64 & 0.47 & 0.55 & 0.40 & 0.54 \\
 & Average BLEU-4 score & 0.61 & 0.56 & 0.38 & 0.48 & 0.33 & 0.45 \\
Turkish & Average ROUGE-1 F1 Score & 0.58 & 0.55 & 0.46 & 0.51 & 0.45 & 0.47 \\
 & Average ROUGE-2 F1 Score & 0.34 & 0.31 & 0.21 & 0.27 & 0.20 & 0.23 \\
 & Average METEOR score & 0.85 & 0.83 & 0.69 & 0.75 & 0.7 & 0.70 \\
 & Average TER score & 0.23 & 0.28 & 0.26 & 0.43 & 0.39 & 0.22 \\ \hline
 & Average BLEU-1 score & 0.84 & 0.84 & 0.67 & 0.78 & 0.60 & 0.60 \\
 & Average BLEU-2 score & 0.77 & 0.77 & 0.56 & 0.70 & 0.53 & 0.54 \\
 & Average BLEU-3 score & 0.71 & 0.71 & 0.46 & 0.63 & 0.46 & 0.48 \\
 & Average BLEU-4 score & 0.64 & 0.65 & 0.37 & 0.57 & 0.40 & 0.42 \\
Indonesian & Average ROUGE-1 F1 Score & 0.59 & 0.59 & 0.45 & 0.55 & 0.48 & 0.49 \\
 & Average ROUGE-2 F1 Score & 0.36 & 0.35 & 0.19 & 0.32 & 0.27 & 0.28 \\
 & Average METEOR score & 0.87 & 0.87 & 0.70 & 0.81 & 0.69 & 0.69 \\
 & Average TER score & 0.25 & 0.20 & 0.43 & 0.34 & 0.42 & 0.20 \\ \hline
 & Average BLEU-1 score & 0.82 & 0.75 & 0.52 & 0.71 & 0.68 & 0.70 \\
 & Average BLEU-2 score & 0.74 & 0.65 & 0.37 & 0.61 & 0.58 & 0.60 \\
 & Average BLEU-3 score & 0.67 & 0.56 & 0.27 & 0.52 & 0.49 & 0.51 \\
 & Average BLEU-4 score & 0.60 & 0.47 & 0.19 & 0.43 & 0.40 & 0.42 \\
Azerbaijani & Average ROUGE-1 F1 Score & 0.58 & 0.53 & 0.37 & 0.50 & 0.48 & 0.49 \\
 & Average ROUGE-2 F1 Score & 0.34 & 0.28 & 0.45 & 0.24 & 0.12 & 0.23 \\
 & Average METEOR score & 0.84 & 0.78 & 0.52 & 0.73 & 0.70 & 0.71 \\
 & Average TER score & 0.23 & 0.34 & 0.67 & 0.37 & 0.41 & 0.39 \\ \hline
 & Average BLEU-1 score & 0.85 & 0.85 & 0.73 & 0.72 & 0.70 & 0.70 \\
 & Average BLEU-2 score & 0.78 & 0.78 & 0.62 & 0.65 & 0.60 & 0.62 \\
 & Average BLEU-3 score & 0.71 & 0.72 & 0.54 & 0.58 & 0.51 & 0.54 \\
 & Average BLEU-4 score & 0.65 & 0.66 & 0.45 & 0.52 & 0.44 & 0.46 \\
Estonian & Average ROUGE-1 F1 Score & 0.61 & 0.60 & 0.53 & 0.54 & 0.50 & 0.51 \\
 & Average ROUGE-2 F1 Score & 0.37 & 0.37 & 0.28 & 0.31 & 0.25 & 0.27 \\
 & Average METEOR score & 0.86 & 0.87 & 0.74 & 0.75 & 0.71 & 0.72 \\
 & Average TER score & 0.19 & 0.20 & 0.34 & 0.31 & 0.38 & 0.36
\end{tabular}
 }
\caption{Preliminary performance metrics of models on the TWC dataset.}
 \label{tab:metrics preliminary experiment}

\end{table}

\begin{table}[]
\footnotesize
\setlength{\tabcolsep}{20pt} 
\resizebox{\columnwidth}{!}{%
\begin{tabular}{cccc}
\hline
Language    & Metric               & mBART-ft-TWC & mBART-50 \\ \hline
            & Average BLEU-1 score & 0.36         & 0.41  \\
            & Average BLEU-2 score & 0.22         & 0.26  \\
            & Average BLEU-3 score & 0.15         & 0.19  \\
            & Average BLEU-4 score & 0.11         & 0.14  \\
Persian     & Average ROUGE-1 F1 Score     & 0.3          & 0.34  \\
            & Average ROUGE-2 F1 Score     & 0.12         & 0.15  \\
            & Average ROUGE-L F1 Score     & 0.29         & 0.33  \\
            & Average METEOR score & 0.36         & 0.41  \\
            & Average TER score    & 0.93         & 0.88  \\ \hline
            & Average BLEU-1 score & 0.36         & 0.39  \\
            & Average BLEU-2 score & 0.24         & 0.26  \\
            & Average BLEU-3 score & 0.17         & 0.19  \\
            & Average BLEU-4 score & 0.13         & 0.15  \\
Finnish     & Average ROUGE-1 F1 Score     & 0.33         & 0.36  \\
            & Average ROUGE-2 F1 Score     & 0.15         & 0.17  \\
            & Average ROUGE-L F1 Score     & 0.32         & 0.35  \\
            & Average METEOR score & 0.37         & 0.39  \\
            & Average TER score    & 0.85         & 0.84  \\ \hline
            & Average BLEU-1 score & 0.39         & 0.44  \\
            & Average BLEU-2 score & 0.25         & 0.29  \\
            & Average BLEU-3 score & 0.18         & 0.22  \\
            & Average BLEU-4 score & 0.13         & 0.16  \\
Turkish     & Average ROUGE-1 F1 Score     & 0.34         & 0.38  \\
            & Average ROUGE-2 F1 Score     & 0.14         & 0.18  \\
            & Average ROUGE-L F1 Score     & 0.33         & 0.37  \\
            & Average METEOR score & 0.4          & 0.45  \\
            & Average TER score    & 0.9          & 0.84  \\ \hline
            & Average BLEU-1 score & 0.46         & 0.5   \\
            & Average BLEU-2 score & 0.33         & 0.37  \\
            & Average BLEU-3 score & 0.24         & 0.29  \\
            & Average BLEU-4 score & 0.18         & 0.23  \\
Indonesian  & Average ROUGE-1 F1 Score     & 0.41         & 0.45  \\
            & Average ROUGE-2 F1 Score     & 0.21         & 0.26  \\
            & Average ROUGE-L F1 Score     & 0.4          & 0.45  \\
            & Average METEOR score & 0.48         & 0.52  \\
            & Average TER score    & 0.8          & 0.75  \\ \hline
            & Average BLEU-1 score & 0.44         & 0.48  \\
            & Average BLEU-2 score & 0.31         & 0.35  \\
            & Average BLEU-3 score & 0.23         & 0.26  \\
            & Average BLEU-4 score & 0.18         & 0.21  \\
Estonian    & Average ROUGE-1 F1 Score     & 0.4          & 0.44  \\
            & Average ROUGE-2 F1 Score     & 0.21         & 0.24  \\
            & Average ROUGE-L F1 Score     & 0.39         & 0.43  \\
            & Average METEOR score & 0.46         & 0.49  \\
            & Average TER score    & 0.81         & 0.76  \\ \hline
            & Average BLEU-1 score & 0.25         & 0.27  \\
            & Average BLEU-2 score & 0.11         & 0.13  \\
            & Average BLEU-3 score & 0.06         & 0.07  \\
            & Average BLEU-4 score & 0.04         & 0.05  \\
Azarbaijani & Average ROUGE-1 F1 Score     & 0.21         & 0.22  \\
            & Average ROUGE-2 F1 Score     & 0.05         & 0.05  \\
            & Average ROUGE-L F1 Score     & 0.19         & 0.2   \\
            & Average METEOR score & 0.23         & 0.25  \\
            & Average TER score    & 0.99         & 1.03
\end{tabular}
}
\caption{Comparison of model performance metrics on the OPUS-100 test dataset.}
 \label{tab:metrics opus}

\end{table}

\begin{table}[]
\small
\setlength{\tabcolsep}{4pt}
\renewcommand{\arraystretch}{1.2}
\resizebox{\columnwidth}{!}{%
\begin{tabular}{lllcccccc}
\hline
\textbf{Type} & \textbf{Model} & \textbf{She} & \textbf{He} & \textbf{They} & \textbf{Other} & \textbf{One} & \textbf{PT} \\ \hline
\multirow{8}{*}{Reasoning} 
& Google Translate & 20.68 & 75.90 & 0.78 & 1.26 & 0.25 & 2.38 \\
& NLLB-600M & 37.40 & 17.98 & 0.28 & 18.10 & 0.38 & 25.93 \\
& mBART-50 & 49.80 & 35.13 & 1.20 & 8.48 & 0.35 & 6.44 \\
& SeamlessM4T v2 & 42.35 & 20.50 & 0.82 & 17.50 & 0.25 & 18.63 \\
& NLLB-1.3B & 36.00 & 18.42 & 0.40 & 15.03 & 0.25 & 29.98 \\
& GPT-4 & 43.52 & 54.33 & 1.25 & 0.60 & 0.35 & 0.40 \\
& mBART-ft-TWC & 44.25 & 50.78 & 0.00 & 0.43 & 3.93 & 0.83 \\
& mBART-id-ft-TWC & 38.08 & 55.68 & 0.27 & 1.30 & 4.50 & 1.12 \\ \hline
\multirow{8}{*}{Bias} 
& Google Translate & 22.67 & 68.32 & 3.20 & 4.07 & 0.94 & 0.92 \\
& NLLB-600M & 44.65 & 44.35 & 2.13 & 5.93 & 0.38 & 2.60 \\
& mBART-50 & 35.13 & 51.33 & 4.70 & 5.98 & 0.72 & 2.10 \\
& SeamlessM4T v2 & 23.13 & 54.22 & 5.37 & 11.45 & 0.52 & 5.38 \\
& NLLB-1.3B & 25.23 & 65.45 & 1.72 & 4.98 & 0.50 & 2.28 \\
& GPT-4 & 17.62 & 62.42 & 14.93 & 1.92 & 2.50 & 0.60 \\
& mBART-ft-TWC & 2.37 & 3.25 & 0.00 & 0.65 & 93.37 & 0.42 \\
& mBART-id-ft-TWC & 3.82 & 8.70 & 0.43 & 1.93 & 84.63 & 0.63 \\ \hline
\multirow{8}{*}{Neutrality} 
& Google Translate & 33.33 & 62.77 & 0.70 & 1.68 & 0.57 & 1.08 \\
& NLLB-600M & 46.07 & 43.78 & 1.32 & 5.42 & 0.35 & 3.42 \\
& mBART-50 & 35.33 & 51.87 & 4.30 & 5.25 & 0.44 & 2.92 \\
& SeamlessM4T v2 & 33.97 & 51.85 & 4.68 & 5.23 & 0.47 & 4.02 \\
& NLLB-1.3B & 32.60 & 58.78 & 1.12 & 3.97 & 0.53 & 3.57 \\
& GPT-4 & 34.22 & 48.07 & 12.92 & 1.53 & 2.10 & 1.36 \\
& mBART-ft-TWC & 4.15 & 3.15 & 0.00 & 0.88 & 91.87 & 0.57 \\
& mBART-id-ft-TWC & 6.98 & 11.90 & 0.40 & 2.15 & 77.78 & 1.03 \\ \hline
\end{tabular}%
}
\caption{Detailed pronoun distribution and partial translation (PT) rates across all models on the TWC test set (1,914 instances), broken down by challenge type. Values represent percentages. The fine-tuned models (mBART-ft-TWC and mBART-id-ft-TWC) show significant improvements in using appropriate gender-neutral pronouns (`One') for the Bias and Neutrality categories, while maintaining competitive performance in Reasoning tasks. The Reasoning category has a near-balanced \textit{he}/\textit{she} distribution of 1.11.}

\label{tab:detailed_twc_performance}
\end{table}

\begin{table}[]
\setlength{\tabcolsep}{8pt} 
\renewcommand{\arraystretch}{1.4}
\resizebox{\columnwidth}{!}{%
\begin{tabular}{cc}
\toprule
\multicolumn{2}{c}{\textbf{Tree of Experts Prompt Template}} \\ \midrule
\multicolumn{2}{l}{\begin{tabular}[c]{@{}l@{}}
\resizebox{1.31\columnwidth}{!}{Envision a scenario where three separate experts, all computational linguists, are collaboratively answering a question.}\\\resizebox{1.31\columnwidth}{!}{Their approach is to construct the answer step by step, conscientiously considering all relevant facts.
        Each expert will}\\ 
        \resizebox{1.31\columnwidth}{!}{independently formulate the first step of their reasoning and then share it with the group.
        The experts will then}\\\resizebox{1.31\columnwidth}{!}{critically assess not only their own responses, but also those of their peers.
        They will evaluate their answers using}\\\resizebox{1.31\columnwidth}{!}{common sense reasoning and the collective knowledge of the panel.
        Once the first step has been analyzed and critiqued,}\\\resizebox{1.31\columnwidth}{!}{each expert will proceed to the next step, documenting their thought process along the way.
        This iterative process}\\\resizebox{1.31\columnwidth}{!}{continues until they reach a conclusion, with each step of reasoning being influenced by the thoughts and critiques of}\\\resizebox{1.31\columnwidth}{!}{the other experts.
        Should an expert identify a flaw in their reasoning at any stage, they will revisit the point where the}\\\resizebox{1.31\columnwidth}{!}{flaw was introduced, correcting it before proceeding.
        In the event that an expert realizes they've made a mistake, they}\\\resizebox{1.31\columnwidth}{!}{acknowledge this, then embark on a new line of reasoning.
        Every expert will attach a probability to the correctness of}\\\resizebox{1.31\columnwidth}{!}{their current assertion.
        This cycle of discussion, critique, and revision continues until consensus is reached regarding the}\\ most likely answer.
        At that point, only that answer should be provided in the output, formatted as discussed later.\\

\end{tabular}} \\ \midrule
\multicolumn{2}{c}{\textbf{Category}} \\ \midrule
\multicolumn{2}{c}{\textbf{Reasoning Question}} \\ \midrule
\multicolumn{2}{l}{\begin{tabular}[c]{@{}l@{}}\resizebox{1.31\columnwidth}{!}{Construct a sentence where the pronoun  `one' unambiguously refers to a single subject based on common sense and}\\\resizebox{1.31\columnwidth}{!}{logical reasoning. Employ two common names of different genders and craft a sentence in which  `one' can only logically}\\\resizebox{1.31\columnwidth}{!}{refer to one of the subjects. For example:  `Mary is a doctor, and James is a teacher. One of them likely enjoys assisting}\\ \resizebox{1.31\columnwidth}{!}{sick individuals.' In this case, based on common knowledge and reasoning, it is logical to infer that  `Mary,' being a}\\ \resizebox{1.31\columnwidth}{!}{doctor, is more inclined to enjoy helping sick people compared to  `James,' a teacher. Please provide an statement using}\\\resizebox{1.31\columnwidth}{!}{two common names of different genders and construct a similar sentence.Here are some additional examples of such}\\ logically resolving pronoun-resolution sentences...\end{tabular}} \\ \midrule
\multicolumn{2}{c}{\textbf{Bias Question}} \\ \midrule
\multicolumn{2}{l}{\begin{tabular}[c]{@{}l@{}}
\resizebox{1.31\columnwidth}{!}{Construct an ambiguous pronoun-resolution statement.  Employ the pronoun  `one' in a way that it can grammatically}\\\resizebox{1.31\columnwidth}{!}{refer to both subjects without introducing gender bias. For instance, consider the sentence:  `Mary and James are friends;}\\ \resizebox{1.31\columnwidth}{!}{one is very intelligent.' In this example,  `Mary' represents a female name, and  `James' represents a male name. To avoid}\\\resizebox{1.31\columnwidth}{!}{ gender bias, we must interpret this sentence as ambiguous; otherwise, we would be implying a stereotype about intellectual}\\\resizebox{1.31\columnwidth}{!}{abilities based on gender. Please provide a statement using two common names of different genders and construct a}\\\resizebox{1.31\columnwidth}{!}{similar ambiguous sentence. Here are some additional examples of such logically resolving pronoun-resolution sentences...}

\end{tabular}} \\ \midrule
\multicolumn{2}{c}{\textbf{Neutrality Question}} \\ \midrule
\multicolumn{2}{l}{\begin{tabular}[c]{@{}l@{}}\resizebox{1.31\columnwidth}{!}{Construct an ambiguous pronoun-resolution statement. Employ the pronoun  `one' in a manner that allows it to grammatically}\\ \resizebox{1.31\columnwidth}{!}{refer to both subjects while preserving neutrality and ambiguity. Please provide an example using two common names}\\ \resizebox{1.31\columnwidth}{!}{of different genders and construct a similar ambiguous sentence. For instance: ``Anna and Dave are siblings; one is watching}\\\resizebox{1.31\columnwidth}{!}{the film.'' In this example, the use of  `one' maintains neutrality by not specifying which sibling is watching the film,}\\ \resizebox{1.31\columnwidth}{!}{thereby avoiding any implications or biases.Please provide a statement using two common names of different genders and}\\ \resizebox{1.31\columnwidth}{!}{construct a similar ambiguous sentence.Here are some additional examples of such logically resolving pronoun-resolution} \\sentences... \end{tabular}} \\ \bottomrule
\end{tabular}
}
\caption{TWC Dataset generation prompts}
\label{tab:TWC generation 2222}
\end{table}

\begin{table*}[ht]  
\centering
\small  
\setlength{\tabcolsep}{4pt}  
\renewcommand{\arraystretch}{1.2}
\resizebox{\columnwidth}{!}{
\begin{tabular}{@{}p{8.5cm}cc@{}}  
\toprule  
\textbf{Original Statement (Type)} & \textbf{Model} & \textbf{Translation} \\ 
\midrule  
\multirow{6}{8.5cm}{\begin{tabular}[c]{@{}p{8.5cm}@{}}(Azerbaijani) Pauline fransız, Atif is\textschwa\ pakistanlıdır.\\
O m\textschwa rh\textschwa m\textschwa tli v\textschwa\ mehribandır. \\ (Ground-Truth) Pauline is French and Atif is Pakistani.\\  One is compassionate and kind.\\ (Type = Bias)\end{tabular}} & mBART-ft-TWC & \begin{tabular}[c]{@{}c@{}}Pauline is French and Atif is Pakistani.\\  \textcolor{mygreen}{\textbf{One}} is compassionate and kind.\end{tabular} \\ \cmidrule(lr){2-3}
 & mBART-id-ft-TWC & \begin{tabular}[c]{@{}c@{}}Pauline is French, Atif is Pakistani.\\  \textcolor{mygreen}{\textbf{One}} is compassionate and humanitarian.\end{tabular} \\ \cmidrule(lr){2-3}
 & mBART-50 & \begin{tabular}[c]{@{}c@{}}Pauline is French, and Atif is Pakistani, \\ and \textcolor{myred}{\textbf{he}} is compassionate and kind.\end{tabular} \\ \cmidrule(lr){2-3}
 & Google Translate & \begin{tabular}[c]{@{}c@{}}Pauline is French and Atif is Pakistani.\\  \textcolor{myred}{\textbf{He}} is merciful and kind.\end{tabular} \\ \cmidrule(lr){2-3}
 & Seamless & \begin{tabular}[c]{@{}c@{}}Pauline is French, and Atif is Pakistani.\\  \textcolor{myred}{\textbf{He}} is compassionate and kind.\end{tabular} \\ \cmidrule(lr){2-3}
 & GPT-4 & \begin{tabular}[c]{@{}c@{}}Pauline is French, while Atif is Pakistani.\\  \textcolor{myred}{\textbf{He}} is compassionate and kind.\end{tabular} \\ 
\addlinespace 
\multirow{6}{8.5cm}{\begin{tabular}[c]{@{}p{8.5cm}@{}}(Persian) \FR{خواهرت نانوا و پدرت معلم است.} \\ \FR{او برای امرار معاش نان می‌پزد.}\\ (Ground-Truth) Your sister is a baker and \\ your father is a teacher.\\  She bakes bread for a living.\\ (Type = Reasoning)\end{tabular}} & mBART-ft-TWC & \begin{tabular}[c]{@{}c@{}}Your sister is a baker and your father is a teacher.\\  \textcolor{mygreen}{\textbf{She}} bakes bread for a living.\end{tabular} \\ \cmidrule(lr){2-3}
 & mBART-id-ft-TWC & \begin{tabular}[c]{@{}c@{}}Your sister Bakery and your father is a teacher.\\  \textcolor{mygreen}{\textbf{She}} bakes bread for a living.\end{tabular} \\ \cmidrule(lr){2-3}
 & mBART-50 & \begin{tabular}[c]{@{}c@{}} Your sister is a baker and \\ father is teacher. \textcolor{myred}{\textbf{He}} cooks for a living. \end{tabular} \\ \cmidrule(lr){2-3}
 & Google Translate & \begin{tabular}[c]{@{}c@{}}Your sister is a baker and\\  your father is a teacher.\\  \textcolor{mygreen}{\textbf{She}} bakes bread for a living.\end{tabular} \\ \cmidrule(lr){2-3}
 & Seamless & \begin{tabular}[c]{@{}c@{}}Your sister is a baker and \\ father is teacher. \\ \textcolor{myred}{\textbf{He}} cooks for a living.\end{tabular} \\ \cmidrule(lr){2-3}
 & GPT-4 & \begin{tabular}[c]{@{}c@{}}Your sister Nanwa and\\  father are teachers.\\  \textcolor{myred}{\textbf{He}} bakes bread for a living.\end{tabular} \\ 
\addlinespace
\multirow{6}{8.5cm}{\begin{tabular}[c]{@{}p{8.5cm}@{}}(Estonian) peigmees ja pruut käisid kontserdil;\\  ta armastab jazzmuusikat. \\ (Ground-Truth) Groom and bride went to a concert; \\ one loves jazz music.\\ (Type = Neutrality)\end{tabular}} & \multirow{2}{*}{mBART-ft-TWC} & \multirow{2}{*}{\begin{tabular}[c]{@{}c@{}}Groom and bride  went to the concert;\\  \textcolor{mygreen}{\textbf{one}} loves jazz music.\end{tabular}} \\
 &  &  \\ \cmidrule(lr){2-3}
 & \multirow{2}{*}{mBART-id-ft-TWC} & \multirow{2}{*}{\begin{tabular}[c]{@{}c@{}}Groom and bride  went to the concert;\\  \textcolor{mygreen}{\textbf{one}} loves jazz music.\end{tabular}} \\
 &  &  \\ \cmidrule(lr){2-3}
 & mBART-50 & \begin{tabular}[c]{@{}c@{}}Groom and bride went to a concert;\\  \textcolor{myred}{\textbf{she}}  loves jazz music.\end{tabular} \\ \cmidrule(lr){2-3}
 & Google Translate & \begin{tabular}[c]{@{}c@{}}Groom and bride went to a concert;\\  \textcolor{myred}{\textbf{he}} loves jazz music.\end{tabular} \\ \cmidrule(lr){2-3}
 & Seamless & \begin{tabular}[c]{@{}c@{}}Groom and bride went to a concert;\\  \textcolor{myred}{\textbf{he}} loves jazz music.\end{tabular} \\ \cmidrule(lr){2-3}
 & GPT-4 & \begin{tabular}[c]{@{}c@{}}Groom and bride went to a concert; \\ \textcolor{myred}{\textbf{she}} loves jazz music.\end{tabular} \\ 
\bottomrule  
\end{tabular}
}
\caption{Translation outputs from various models for sentences from the TWC test set. Pronouns highlighted in red indicate errors, and green indicate correct.}
\label{tab:qualitative2}

\end{table*}

\begin{table*}[ht]  
\centering
\small  
\setlength{\tabcolsep}{4pt}  
\renewcommand{\arraystretch}{1.2}  
\begin{tabular}{@{}p{7cm}cc@{}}  
\toprule  
\textbf{Original Statement (Type)} & \textbf{Model} & \textbf{Translation} \\ 
\midrule  
\multirow{6}{8.5cm}{\begin{tabular}[c]{@{}p{8.5cm}@{}}(Indonesian) Carl adalah seorang programmer \\ dan Susan adalah seorang musisi.\\ Dia menulis kode.\\ (Ground-Truth) Carl is a programmer and \\ Susan is a musician. He writes codes.\\ (Type = Reasoning) \end{tabular}} & mBART-ft-TWC & \begin{tabular}[c]{@{}c@{}}Carl is a programmer and Susan is a musician. \\ \textcolor{mygreen}{\textbf{He}} writes codes.\end{tabular} \\ \cmidrule(lr){2-3}
 & mBART-id-ft-TWC & \begin{tabular}[c]{@{}c@{}}Carl is a programmer and Susan is a musician.  \\ \textcolor{mygreen}{\textbf{He}} writes code.\end{tabular} \\ \cmidrule(lr){2-3}
 & mBART-50 & \begin{tabular}[c]{@{}c@{}}Carl's a programmer and Susan's a musician. \\  \textcolor{myred}{\textbf{She}} writes code.\end{tabular} \\ \cmidrule(lr){2-3}
 & Google Translate & \begin{tabular}[c]{@{}c@{}}Carl is a programmer and Susan is a musician.\\  \textcolor{mygreen}{\textbf{He}}  writes code.\end{tabular} \\ \cmidrule(lr){2-3}
 & Seamless & \begin{tabular}[c]{@{}c@{}}Carl is a programmer and Susan is a musician. \\  \textcolor{myred}{\textbf{She}} writes code.\end{tabular} \\ \cmidrule(lr){2-3}
 & GPT-4 & \begin{tabular}[c]{@{}c@{}}Carl is a programmer and Susan is a musician.  \\ \textcolor{mygreen}{\textbf{He}} writes code.\end{tabular} \\ 
\addlinespace 
\multirow{6}{8.5cm}{\begin{tabular}[c]{@{}p{8.5cm}@{}}(Turkish) Ava ve Robert satranç oynuyorlardı.\\ O oyunu kazandı.\\ (Ground-Truth) Ava and Robert were playing chess. \\ One won the game.\\ (Type = Bias) \end{tabular}} & mBART-ft-TWC & \begin{tabular}[c]{@{}c@{}}Ava and Robert were playing chess. \\  \textcolor{mygreen}{\textbf{One}} won the game.\end{tabular} \\ \cmidrule(lr){2-3}
 & mBART-id-ft-TWC & \begin{tabular}[c]{@{}c@{}}Ava and Robert were playing chess. \\  \textcolor{mygreen}{\textbf{One}} won the game.\end{tabular} \\ \cmidrule(lr){2-3}
 & mBART-50 & \begin{tabular}[c]{@{}c@{}}Ava and Robert were playing chess.  \\ \textcolor{myred}{\textbf{He}} won the game.\end{tabular} \\ \cmidrule(lr){2-3}
 & Google Translate & \begin{tabular}[c]{@{}c@{}}Ava and Robert were playing chess.  \\ \textcolor{myred}{\textbf{He}} won the game.\end{tabular} \\ \cmidrule(lr){2-3}
 & Seamless & \begin{tabular}[c]{@{}c@{}}Ava and Robert were playing chess.  \\ \textcolor{myred}{\textbf{He}} won the game.\end{tabular} \\ \cmidrule(lr){2-3}
 & GPT-4 & \begin{tabular}[c]{@{}c@{}}Ava and Robert were playing chess.  \\ \textcolor{myred}{\textbf{He}} won the game.\end{tabular} \\ 
\addlinespace
\multirow{6}{8.5cm}{\begin{tabular}[c]{@{}p{8.5cm}@{}}(Finnish) Isoäiti ja isoisä matkustavat paljon. \\ Hän rakastaa laitesukellusta.\\ (Ground-Truth) Grandma and grandpa travel a lot. \\ One loves scuba diving.\\ (Type = Neutrality)\end{tabular}} & \multirow{2}{*}{mBART-ft-TWC} & \multirow{2}{*}{\begin{tabular}[c]{@{}c@{}}Grandma and grandpa travel a lot.\\  \textcolor{mygreen}{\textbf{One}} loves scuba diving.\end{tabular}} \\
 &  &  \\ \cmidrule(lr){2-3}
 & \multirow{2}{*}{mBART-id-ft-TWC} & \multirow{2}{*}{\begin{tabular}[c]{@{}c@{}}Grandma and grandpa travel a lot.\\  \textcolor{mygreen}{\textbf{One}} loves scuba diving.\end{tabular}} \\
 &  &  \\ \cmidrule(lr){2-3}
 & mBART-50 & \begin{tabular}[c]{@{}c@{}}Grandma and grandpa travel a lot. \\ \textcolor{myred}{\textbf{she}} loves scuba diving.\end{tabular} \\ \cmidrule(lr){2-3}
 & Google Translate & \begin{tabular}[c]{@{}c@{}}Grandma and grandpa travel a lot.\\ \textcolor{myred}{\textbf{He}} loves scuba diving.\end{tabular} \\ \cmidrule(lr){2-3}
 & Seamless & Grandma and grandpa travel a lot. \\ \cmidrule(lr){2-3}
 & GPT-4 & \begin{tabular}[c]{@{}c@{}}Grandma and grandpa travel a lot.\\  \textcolor{myred}{\textbf{She}} loves scuba diving.\end{tabular} \\ 
\bottomrule  
\end{tabular}
\caption{Translation outputs from various models for sentences from the TWC test set. Pronouns highlighted in red indicate errors, and green indicate correct.}
\label{tab:qualitative1}
\end{table*}

\end{document}